\documentclass{article}

\PassOptionsToPackage{numbers, sort&compress}{natbib}

\usepackage[preprint]{neurips_2026}

\usepackage[utf8]{inputenc} %
\usepackage[T1]{fontenc}    %
\usepackage{hyperref}       %
\usepackage{url}            %
\usepackage{booktabs}       %
\usepackage{amsfonts}       %
\usepackage{nicefrac}       %
\usepackage{microtype}      %
\usepackage{xcolor}         %
\usepackage{xspace}
\usepackage{booktabs}
\usepackage{makecell}
\usepackage{pifont}
\usepackage{multirow} 
\usepackage{amsmath}
\usepackage{graphicx}
\usepackage{caption}
\usepackage[capitalize]{cleveref}
\crefname{section}{Sec.}{Secs.}
\Crefname{section}{Sec.}{Secs.}
\crefname{subsection}{Sec.}{Secs.}
\Crefname{subsection}{Sec.}{Secs.}
\crefname{subsubsection}{Sec.}{Secs.}
\Crefname{subsubsection}{Sec.}{Secs.}
\crefname{figure}{Fig.}{Figs.}
\Crefname{figure}{Fig.}{Figs.}
\crefname{table}{Tab.}{Tabs.}
\Crefname{table}{Tab.}{Tabs.}
\crefname{equation}{Eq.}{Eqs.}
\Crefname{equation}{Eq.}{Eqs.}
\creflabelformat{equation}{#2\textup{#1}#3}

\usepackage{appendix}
\usepackage[utf8]{inputenc}
\usepackage{tcolorbox}
\usepackage{array}
\usepackage{geometry}
\usepackage{enumitem}
\usepackage{algorithm}
\usepackage[noend]{algpseudocode}
\tcbuselibrary{skins, breakable}
\newcommand{\cmark}{\textcolor{green!70!black}{\ding{51}}}
\newcommand{\xmark}{\textcolor{red!40}{\ding{55}}}

\newcommand{\cmarkb}{\textcolor{black!90}{\ding{51}}}
\newcommand{\xmarkb}{\textcolor{black!50}{\ding{55}}}
\newcommand{\ie}{\textit{i.e.}\xspace}
\newcommand{\eg}{\textit{e.g.}\xspace}

\usepackage[subtle]{savetrees}

\definecolor{DeepGreen}{rgb}{0.0, 0.5, 0.0}
\newcommand{\ColorComment}[1]{\Comment{\textcolor{DeepGreen}{#1}}}
\newcommand{\LineCommentBase}[1]{\Statex \textcolor{DeepGreen}{\# #1}}

\usepackage[table]{xcolor} 
\usepackage{colortbl}   
\definecolor{midpurblue}{RGB}{205, 227, 255}
\definecolor{lightpurblue}{RGB}{240, 245, 252}
\newcommand{\samevalue}[1]{\cellcolor{gray!10}#1}
\newcommand{\improv}[1]{\cellcolor{green!10}#1}
\newcommand{\regress}[1]{\cellcolor{red!10}#1}
\newcommand{\method}{\textsc{Gmos}\xspace}
\newcommand{\bestab}[1]{\textbf{#1}}
\newcommand{\secondbestab}[1]{{#1}}
\newcommand{\para}[1]{\vspace{1pt}{\noindent \textbf{#1}}}
\newcommand{\noindentpara}[1]{{\noindent \textbf{#1}}}

\title{\method: Grounding Moving Object Segmentation \\ in 3D Space and Time}

\author{%
  \hspace{0.cm} Junyu~Xie$^1$\hspace{0.3cm} Tengda Han$^{1}$ \hspace{0.25cm}  Weidi Xie$^{1,2}$ \hspace{0.15cm} Andrew Zisserman$^{1}$ \\
  $^1$Visual Geometry Group, Department of Engineering Science, University of Oxford, UK  \\
  $^2$SAI, Shanghai Jiao Tong University, China\\
  \texttt{\{jyx,tengda,weidi,az\}@robots.ox.ac.uk} \\
  {\url{https://www.robots.ox.ac.uk/vgg/research/gmos/}}
}

\begin{document}

\maketitle \vspace{-2mm}

\begin{figure*}[h]
\includegraphics[width=1.0\linewidth]{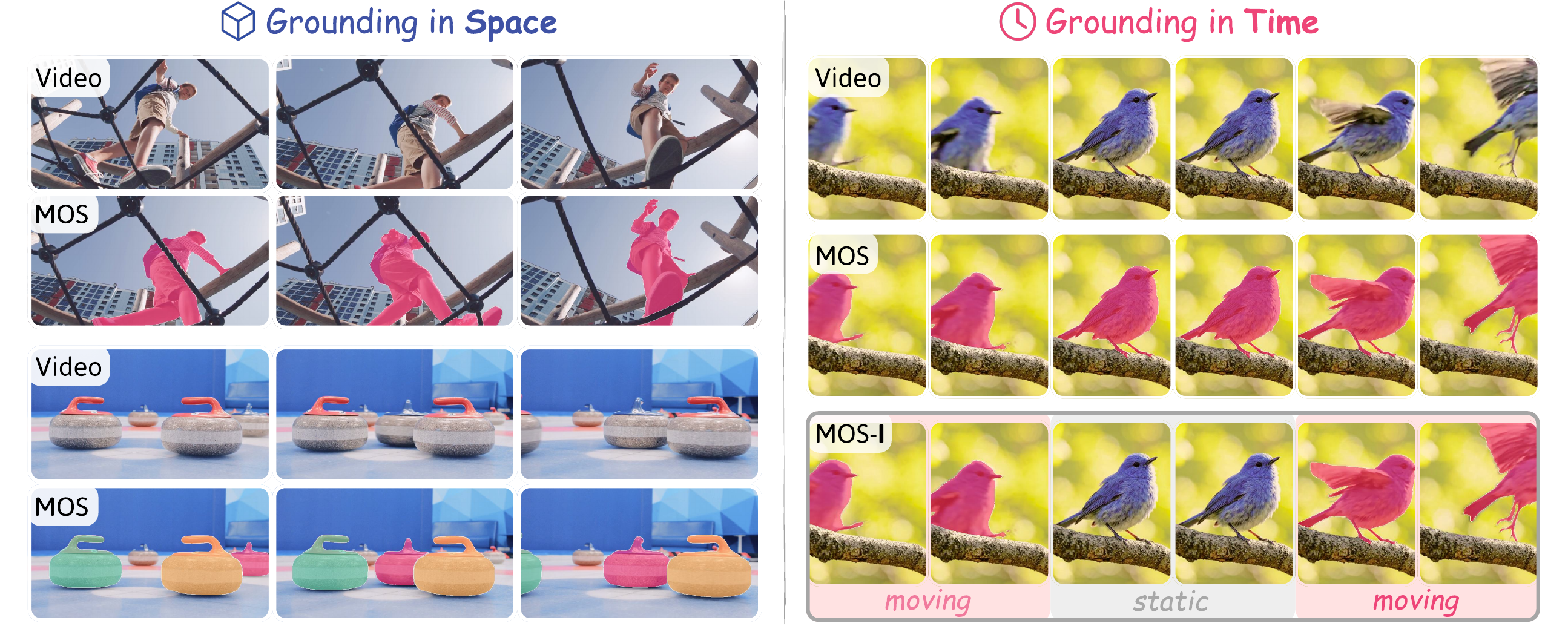}
\vspace{-4.mm}
\caption{
\textbf{Grounding Moving Object Segmentation (MOS) in 3D space and time.}
\textbf{Left:} \method grounds MOS in 3D \textit{space}, enabling reliable segmentation under challenging viewpoints and heavy depth parallax from moving cameras.
\textbf{Right:} \method also grounds MOS in \textit{time}: conventional MOS masks every object that moves at some point in the sequence, regardless of whether it is currently in motion. In contrast, our proposed \textit{MOS-I} (``I'' for Instantaneous) protocol requires predicting, for every frame, which objects are \emph{actively moving} at that instant. This is illustrated in the bottom row, where a small bird lands, stops, then takes off, while the tree branch remains static throughout.
}
\vspace{2mm}
\label{fig:teaser}
\end{figure*}

\begin{abstract}
Moving Object Segmentation (MOS) aims to discover, segment, and track objects that move independently of the camera. Current MOS methods, however, exhibit two fundamental limitations: they rely on pre-computed 2D auxiliary modalities such as optical flow or point trajectories that lack 3D geometric information, and they treat motion as a sequence-level attribute, overlooking the instantaneous motion state of each object. We address both by grounding MOS in 3D space and time, and propose \method, a framework that operates directly on RGB video to produce 3D-aware, temporally fine-grained segmentation of multiple moving objects, alongside a foreground--background variant \method-S for faster deployment. To support training and evaluation in this regime, we curate \method-2K, a dataset of $2{,}210$ real-world videos with per-object temporal motion annotations drawn from five established Video Object Segmentation (VOS) benchmarks, and formalise MOS-I (``I'' for {instantaneous}), a temporally fine-grained evaluation protocol with three complementary metrics. \method achieves state-of-the-art results across MOS, MOS-I, and Unsupervised VOS benchmarks, while running significantly faster than prior multi-object MOS methods and supporting online inference for streaming deployment.
\end{abstract}

\section{Introduction}
\label{sec:introduction}

Moving object segmentation (MOS) aims to discover, segment, and track objects that move independently of the camera. This capability is foundational to applications such as autonomous driving~\cite{chen2021lidarmos, li2023mosfusion}, camouflaged object detection~\cite{Lamdouar20, Yang21a}, and video surveillance~\cite{Nasaruddin2020, weng2021aten}, and has become increasingly central to recent 3D/4D scene reconstruction methods~\cite{Li_2025_CVPR, golisabour2025romo} that require motion masks to separate dynamic objects from the static scene.

Current MOS methods exhibit two fundamental limitations, as outlined in~\Cref{tab:method-comparison}. \emph{First}, they are not 3D-aware and remain fragile under large camera motion or depth parallax. This stems from their heavy reliance on pre-computed 2D auxiliary modalities (\eg optical flow~\cite{Yang21a, Xie22, Lian_2023_CVPR} or point trajectories~\cite{Huang_2025_CVPR, Karazija24b}), which lack 3D geometric information and incur substantial pre-processing overhead. \emph{Second}, existing benchmarks assume that objects move throughout the video, leading methods to treat motion as a \emph{sequence-level} attribute and to overlook the instantaneous motion state of each object. This coarse temporal granularity precludes online and real-time deployment.

We address both limitations by grounding MOS in \emph{space} and \emph{time}. Our approach rests on two observations: (\textit{i})~the geometric representations learnt by recent feed-forward 3D reconstruction models already encode sufficient information to disentangle camera and object motion, and (\textit{ii})~independent object motion is reliably identifiable from a short temporal window ($\sim$$0.5$\,s).

Building on these insights, we propose \method, a proposer--propagator framework that operates directly on RGB video without any auxiliary pre-computed modalities. A per-frame \emph{proposer} fuses geometric features from a $\pi^3$ encoder~\cite{wang2026pi} with segmentation features from a SAM2 encoder~\cite{ravi2024sam2}, jointly producing object mask proposals and per-frame motion-state predictions. A SAM2-based \emph{propagator} then links these proposals into consistent tracks across frames. We further introduce \method-S (``S'' for \emph{single}), a streaming foreground--background variant that predicts all moving objects as a \emph{single binary} mask and dispenses with the propagator for even faster inference. 

\begin{table}[t]
\centering
\small
\caption{Comparison with Moving Object Segmentation (MOS) baselines. *Auxiliary inputs typically involve optical flow, 2D tracks and depths. The runtime measurement setting is detailed in Appendix~\ref{appsec:imple}.}
\vspace{1.5mm}
\label{tab:method-comparison}
\setlength{\tabcolsep}{4pt}
\resizebox{0.9\linewidth}{!}{%
\begin{tabular}{l cccccc}
\toprule
Method & Multi-object & 3D-aware & \makecell{Temporally\\fine-grained} & Online \!\!\!\!\!\!\!\!& \;\;\makecell{No aux. inputs* }\!\!\!\!\!\! & \makecell{Per-frame\\runtime (s)} \\
\midrule
FlowSAM~\cite{Xie24a}    & \cmark & \xmark & \xmark & \xmark\!\!\!\!\!\!\!\! & \xmark\;\textcolor{gray}{(flow)} & $1.79$ \\
SegAnyMo~\cite{Huang_2025_CVPR}  & \cmark & \cmark & \xmark & \xmark\!\!\!\!\!\!\!\! & \;\;\;\;\;\;\;\;\;\;\;\xmark \;\textcolor{gray}{(depth, track)} & $4.06$ \\
GeoMotion~\cite{he2026geomotion} & \xmark & \cmark & \xmark & \xmark\!\!\!\!\!\!\!\! & \xmark\;\textcolor{gray}{(flow)} & $0.87$ \\
\midrule
\method-S    & \xmark & \cmark & \cmark & \cmark\!\!\!\!\!\!\!\! & \cmark & $0.26$ \\
\method      & \cmark & \cmark & \cmark & \cmark\!\!\!\!\!\!\!\! & \cmark & $0.38$ \\
\bottomrule
\end{tabular}}
\vspace{-2mm}
\end{table}

To support MOS in a more geometrically grounded and temporally fine-grained manner, we curate \method-2K, a collection of $2{,}210$ real-world videos drawn from five established VOS benchmarks and annotated with per-frame motion states for every object. The subsets span complementary domains, including heavily occluded scenes, hand--object interactions, and wildlife and human-centric clips. Alongside this dataset, we formalise MOS-I (``I'' for \emph{instantaneous}), a temporally fine-grained MOS evaluation protocol. As illustrated in~\cref{fig:teaser}~(right), MOS-I scores segmentation masks only on frames where objects are actively moving, and penalises false-positive predictions on static and background objects.

\noindent In summary, our contributions are as follows:
\begin{itemize}[leftmargin=1.2em, itemsep=2pt, parsep=2pt, topsep=-1pt]
    \item We propose \method, a proposer--propagator framework that delivers 3D-aware, temporally fine-grained MOS from RGB video alone and supports online inference. We also introduce \method-S, a foreground--background variant for even faster deployment.
    \item We introduce a confidence-based training objective that automatically down-weights noisy frame-level motion annotations near static--moving transitions.
    \item We curate \method-2K, a dataset of $2{,}210$ real-world videos drawn from five established VOS benchmarks and annotated with per-object temporal motion labels, providing a unified resource for both training and evaluating temporally fine-grained MOS methods.
    \item We formalise MOS-I, a temporally fine-grained evaluation protocol for MOS, together with three complementary metrics that jointly measure segmentation accuracy on actively moving frames and robustness to false-positive predictions on static and background objects.
\end{itemize}

\noindent \method achieves state-of-the-art results across MOS, MOS-I, and Unsupervised VOS (UVOS) benchmarks, while running roughly three times faster than prior MOS methods (\Cref{tab:method-comparison}).

\section{Method: \method}
\label{sec:method}

Given an RGB video {\small $\mathcal{V} = \{I_t\}_{t=1}^{T}$}, our goal is to predict, for every frame $t$ and every independently moving object $i$, a segmentation mask together with its motion state,
\begin{equation}
    \bigl\{\, (\hat{M}^{(i)}_t, \hat{m}^{(i)}_t) \,\bigr\}_{t=1,\dots,T;\, i=1,\dots,N} \;=\; \Phi_{\method}(\mathcal{V}),
    \label{eq:task}
\end{equation}
where {\small $\hat{M}^{(i)}_t \in \{0,1\}^{H\times W}$} is a binary segmentation mask, {\small $\hat{m}^{(i)}_t \in \{0,1\}$} is the motion state with {\small $\hat{m}^{(i)}_t = 1$} indicating that object $i$ is moving at frame $t$. \method produces these outputs jointly, and therefore covers both MOS-I (\Cref{subsec:eval}) and the conventional MOS settings.

\Cref{fig:model} (left) provides an overview of \method, which follows a \emph{proposer-propagator} design. The proposer operates on a short local window and outputs, for the middle frame, a set of $N$ per-object proposals, each comprising a moving object mask and a predicted motion state. The propagator then links these per-frame proposals into consistent object tracks across the full video.

\subsection{\method proposer}
\begin{figure*}[t]
\centering
\includegraphics[width=0.95\linewidth]{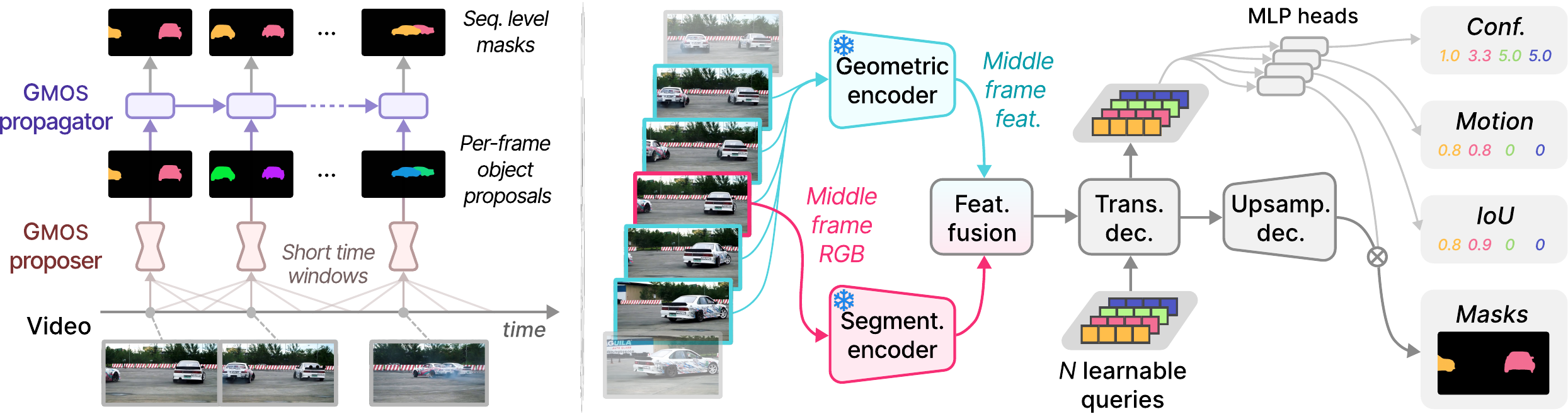}
\vspace{-0.5mm}
\caption{
\textbf{Overview of \method.}
\textbf{Left:} The overall \emph{proposer--propagator} design. The proposer operates on a short temporal window (around $0.5$\,s) and outputs per-frame object proposals, which the propagator links into coherent tracks across the full video.
\textbf{Right:} The \method proposer. A frozen $\pi^3$ geometric encoder ingests frames for a short temporal window and a frozen SAM2 segmentation encoder processes the middle frame of the window, to make predictions for the middle frame. Their features are fused and passed through a transformer decoder with $N$ learnable object queries, decoded into segmentation masks for moving objects alongside motion states, estimated IoUs, and confidence scores.
}
\label{fig:model}
\vspace{-3mm}
\end{figure*}

\label{subsec:proposer}

As illustrated in \Cref{fig:model} (right), the proposer takes raw video frames as its sole input and predicts, for each of $N$ object queries, a segmentation mask together with auxiliary outputs: a confidence score, a motion state, and an estimated mask IoU.

\para{Geometric encoder.} To achieve \emph{3D-awareness},
we extract geometric features using a frozen {\small $\pi^3$} encoder~\cite{wang2026pi}. Rather than feeding the entire sequence, we process a short temporal window {\small  $[t-n,\,t+n]$} centred at the query frame $t$, sufficient for distinguishing independent object motion from camera-induced parallax. After {\small $\pi^3$} encoding, we retain only the \emph{middle-frame} feature {\small $f^{\text{geo}}_t \in \mathbb{R}^{h_{\text{g}} \times w_{\text{g}} \times d_{\text{g}}}$}, as inter-frame correspondence is already encoded in it through the model's cross-frame attention.

\para{Segmentation encoder.} Since geometric features alone encode little about object segmentation cues, we leverage the strong objectness prior of a frozen SAM2 encoder~\cite{ravi2024sam2}. Only the query frame $I_t$ is passed through SAM2, yielding a bottleneck feature {\small $f^{\text{seg}}_t \in \mathbb{R}^{h \times w \times d}$} together with high-resolution features that are later used by the mask head.

\para{Feature fusion.} The geometric feature {\small $f^{\text{geo}}_t$} is typically at a lower spatial resolution than the segmentation one {\small $f^{\text{seg}}_t$}, \ie $h_{\text{g}} < h$. We therefore bilinearly interpolate {\small $f^{\text{geo}}_t$} to $(h, w)$, concatenate the two features channel-wise, and apply a linear projection to obtain the fused feature {\small $f_t \in \mathbb{R}^{h \times w \times d}$}.
A light-weight transformer self-attention encoder is then applied to {\small $f_t$} to enforce global spatial consistency.

\para{Transformer decoder.} Following the convention in query-based segmentation~\cite{cheng2021maskformer, Li_2023_CVPR}, we introduce $N$ learnable object queries {\small $\{q^{(i)}\}_{i=1}^{N}$} with {\small $q^{(i)} \in \mathbb{R}^{d}$} and feed them into a two-way transformer decoder~\cite{kirillov2023segment} together with $f_t$. Each layer alternates query self-attention with query-to-feature and feature-to-query cross-attention, producing refined query tokens {\small $\{q^{*(i)}_t\}$} and refined dense features {\small $f^*_t \in \mathbb{R}^{h \times w \times d}$}.

\para{Mask head.} The refined dense features {\small ${f}^*_t$} are upsampled to full resolution through transposed convolutions, with the SAM2 high-resolution features added as skip connections. An MLP maps each refined query {\small ${q}^{*(i)}_t$} to a mask embedding, and the per-query mask {\small $\hat{M}^{(i)}_t$} is obtained by dot product with the upsampled features, following the hypernetwork design in~\cite{cheng2021maskformer}.
During training, queries are matched to ground truth via bipartite Hungarian assignment. Note, an object's ground-truth mask is defined as \emph{empty} whenever the object is not moving, so supervision jointly shapes the mask and the motion state.

\para{Motion and IoU heads.} Two independent MLP heads take in the refined query {\small ${q}^{*(i)}_t$}. The \emph{motion head} leads to the probability {\small $\hat{m}^{(i)}_t$} that object $i$ is moving at frame $t$, and the \emph{IoU head} produces a scalar {\small $\hat{u}^{(i)}_t$} estimating the quality of {\small $\hat{M}^{(i)}_t$} in terms of its expected IoU against the ground truth.

\para{Confidence prediction.} The instantaneous motion state is inherently ambiguous near transitions between motion and rest, where the ground truth switches abruptly between a full mask with {\small $m^{(i)}_t = 1$} and an empty mask with {\small $m^{(i)}_t = 0$}. To mitigate this label noise, we follow~\cite{Wang_2024_CVPR} and introduce a learnable per-query confidence {\small $\hat{c}^{(i)}_t \in [0, 1]$}, predicted by a final MLP head with sigmoid output. The resulting confidence modulates the mask and motion losses (see~\cref{eq:loss}).

\para{Training objective.} The total proposer loss reads
\begin{equation}
\mathcal{L} = \lambda_{\text{iou}} \mathcal{L}_{\text{iou}}
    + \tilde{c}\bigl( \lambda_{\text{mask}} \mathcal{L}_{\text{mask}} + \lambda_{\text{motion}} \mathcal{L}_{\text{motion}} \bigr)
    - \bigl[ \lambda^{\text{mov}}_{\text{conf}}\;m + \lambda^{\text{sta}}_{\text{conf}}\;(1\!-\!m) \bigr]\,\log\tilde{c}
\label{eq:loss}
\end{equation}
The loss is averaged over object queries $i$ and frames $t$, with the corresponding indices dropped for clarity. Here, $\mathcal{L}_{\text{mask}}$ is a mixture of focal and dice losses, $\mathcal{L}_{\text{motion}}$ is a binary cross-entropy on the motion state, and $\mathcal{L}_{\text{iou}}$ is an MSE loss between the predicted IoU and the actual IoU against the ground truth. 

The mask and motion losses are modulated by a rescaled confidence {\small $\tilde{c} = (c_{\max}-1)\,\hat{c} + 1 \in [1, c_{\max}]$}, and the last term is an adversarial confidence regulariser~\cite{Wang_2024_CVPR} that raises $\tilde{c}$ on clean, easy samples the model can solve precisely. For difficult examples, such as a walking pedestrian coming to a stop, the mask and motion losses are large due to label ambiguity around the transition. The model therefore lowers $\tilde{c}$ to attenuate the multiplicative term, paying a small $\log\tilde{c}$ penalty in exchange.

In practice, the confidence would be driven up only for static samples, since predicting an empty mask for a static object is trivially easier than predicting a precise moving-object mask, failing to address the transition noise. We therefore \emph{split} the regularisation by the ground-truth motion state $m$, using distinct coefficients {\small $\lambda^{\text{mov}}_{\text{conf}}$} and {\small $\lambda^{\text{sta}}_{\text{conf}}$} to calibrate moving and static samples independently.

\para{\method-S: \textit{a foreground--background variant.}}
For efficiency-critical deployment, we additionally instantiate \method-S (``S'' for \emph{single} binary mask prediction), which predicts all moving objects into a single foreground mask. \method-S shares the encoders, feature fusion, and self-attention module with the \method proposer, but (i) removes the query-based transformer decoder, (ii) replaces the per-query mask head with a single convolutional decoder that outputs one binary foreground mask per frame, and (iii) dispenses with the follow-up propagator entirely. This makes \method-S an end-to-end model that produces instantaneous (per-frame) motion segmentation and achieves lower inference latency than \method. Full architectural details are provided in Appendix~\ref{appsubsec:gmos-s}.

\subsection{\method propagator}
\label{subsec:propagator}

For every frame $t$, the proposer yields a set of $N$ candidate tuples {\footnotesize $\{(\hat{M}^{(i)}_t, \hat{m}^{(i)}_t, \hat{u}^{(i)}_t, \hat{c}^{(i)}_t)\}_{i=1}^{N}$}. The role of the propagator is to lift these per-frame proposals {\footnotesize $\{\hat{M}^{(i)}_t\}$} into coherent object tracks across the full video, guided by the predicted motion probabilities {\footnotesize $\{\hat{m}^{(i)}_t\}$} and estimated IoUs {\footnotesize $\{\hat{u}^{(i)}_t\}$}. The confidence scores {\footnotesize $\{\hat{c}^{(i)}_t\}$} are not used here, as they primarily serve as regularisers during proposer training. We build this process on a SAM2 video predictor~\cite{ravi2024sam2} used as a mask-conditioned tracker, and define an elementary propagation step from which we build online and offline procedures.

\para{Basic propagation step.}
Given a start frame and a direction, SAM2 propagates the currently tracked masks frame by frame. At each visited frame $t$, the proposer predictions are first filtered by thresholding on the motion probabilities {\footnotesize $\{\hat{m}^{(i)}_t\}$} and estimated IoUs {\footnotesize $\{\hat{u}^{(i)}_t\}$}. The remaining high-quality moving-object proposals are then matched to the propagated masks via Hungarian assignment based on mask IoU, and the set of active tracks is updated according to one of two cases:
\begin{itemize}[leftmargin=1.2em, itemsep=1pt, parsep=1pt, topsep=-1pt]
    \item \textit{Addition.} If a proposer prediction matches no existing SAM2 track, it is instantiated as a new mask track starting from frame~$t$.
    \item \textit{Reinforcement.} If a proposer prediction matches an existing SAM2 track at high IoU, it is injected as an additional prompt for that track, suppressing drift.
\end{itemize}

For both cases above, the track's motion state at frame~$t$ is set to moving. Conversely, if an existing track has a propagated mask at frame~$t$ but no proposer prediction falls on the same region (\eg the object was moving earlier but has become static), the track's motion state at frame~$t$ is set to static.

\para{Online procedure.}
The online procedure initialises the tracker from the \emph{first frame} and runs the basic propagation steps in a single causal forward pass. This streams a mask and motion state for each frame as soon as it is processed, supporting MOS-I. Since the sequence is viewed only once and causally, masks cannot be recovered for frames preceding an object's first observed motion, making the online procedure incompatible with the conventional MOS setting that requires a mask at every frame.

\para{Offline procedure.}
The offline procedure builds on top of the online results with a second propagation stage. It first selects an \emph{anchor frame} $t^\star$ from the online output, defined as the frame whose moving object proposals carry the highest aggregate predicted IoU. The basic propagation step is then run \emph{bi-directionally}, forward from $t^\star$ to $T$ and backward from $t^\star$ to $1$, yielding a dense mask for every object at every frame. This mode can be used for both MOS and MOS-I. Further algorithmic details and pseudo-code are provided in Appendix~\ref{appsubsec:gmos-prop}.

\section{Dataset: \method-2K}
\label{sec:dataset}
We introduce \method-2K, a video moving object segmentation dataset with temporally fine-grained motion labels. Beyond ground-truth masks for moving objects, it specifies the exact temporal windows during which each object is in motion, enabling precise training and evaluation of time-sensitive motion segmentation methods.

\begin{figure*}[t]
\centering
\includegraphics[width=0.95\linewidth]{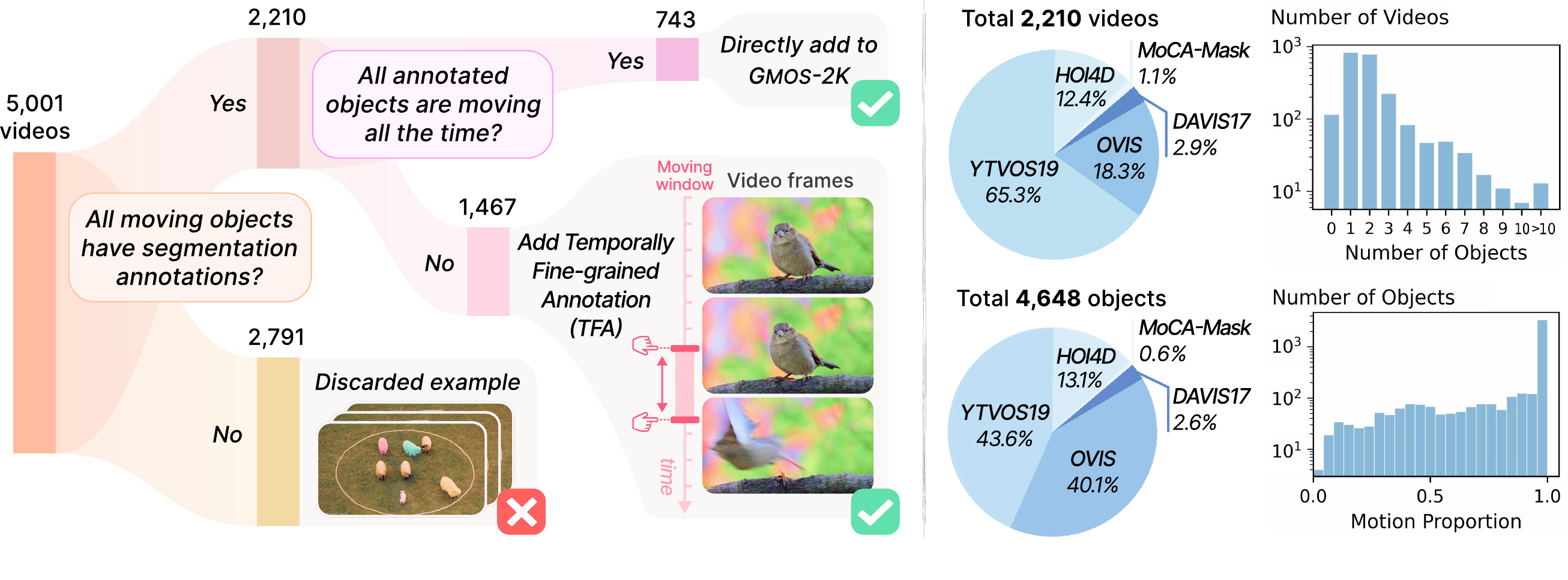}
\vspace{-4.5mm}
\caption{
\textbf{\method-2K overview.} \textbf{Left:} {Curation pipeline.} $5{,}001$ videos from five VOS datasets are filtered by two criteria, with $743$ added directly to \method-2K and $1{,}467$ requiring Temporal Fine-grained Annotation (TFA), which labels per-object motion intervals along the time axis. \textbf{Right:} {Dataset statistics.} Pie charts show the per-subset distribution of videos and objects. The upper histogram reports the number of moving objects per video, and the lower one reports the per-object motion proportion, \ie the fraction of the sequence in which each object is actively moving.}
\label{fig:gmos2k}
\vspace{-4mm}
\end{figure*}

\para{Curation pipeline.}
Temporal motion labels are substantially cheaper to obtain than dense segmentation masks. We therefore build upon five existing video object segmentation datasets (DAVIS17~\cite{Ponttuset17}, YTVOS19~\cite{Xu18}, OVIS~\cite{qi2022occluded}, MoCA-Mask~\cite{Cheng_2022_CVPR}, and HOI4D~\cite{Liu_2022_CVPR}) and annotate them with per-object motion timestamps.
\Cref{fig:gmos2k}~(left) illustrates the data filtering and annotation pipeline. Starting from $5{,}001$ candidate videos, we apply two sequential filters:
\begin{enumerate}[leftmargin=1.6em, itemindent=1pt, itemsep=1pt, parsep=2pt, topsep=-2pt]
    \item[(\textit{i})] \textit{Do all independently moving objects in the video have ground-truth segmentation masks?} \\
    If yes, the sequence proceeds to the second question, otherwise it is filtered out. Note, an object is considered ``moving'' if it undergoes motion independent of the camera for at least one frame. This question concerns only moving-object masks. Static objects without masks are permissible.
    \item[(\textit{ii})] \textit{Are all annotated objects moving throughout the entire sequence?} \\
    If yes, no additional annotation is needed. Otherwise, the video is forwarded for Temporally Fine-grained Annotation (TFA).
\end{enumerate}
\para{Temporally Fine-grained Annotation (TFA).}
Given a video sequence, TFA aims to annotate time intervals on the temporal axis that specify when objects are moving, with each object annotated independently. The annotation process follows that for video temporal grounding. Further details on the annotation interface, guidelines, and quality controls are provided in Appendix~\ref{appsec:gmos_curation}.

\para{Dataset statistics.}
\method-2K comprises $2{,}210$ videos with $4{,}648$ annotated moving objects, drawn from five constituent subsets that span diverse domains: heavily occluded scenes (OVIS), hand-object interactions (HOI4D), and wildlife and human-centric clips (DAVIS17, YTVOS19, MoCA-Mask). 
Example sequences are shown in~\cref{fig:gmos2k_vis}, with instantaneously moving objects labelled. 

The pie charts in \Cref{fig:gmos2k}~(right) show the per-subset distribution of videos and objects. OVIS contributes a disproportionately large share of objects, reflecting its dense multi-object scenes. The accompanying histograms characterise two complementary aspects of the dataset. The per-video object count (top) concentrates on sequences with $0$--$4$ moving objects, with a long tail extending beyond $4$. The motion proportion (bottom), defined as the fraction of frames in which an object is actively moving, peaks at fully moving objects, with the remaining cases distributed roughly uniformly across lower values.

The training and test splits consist of $1{,}930$ and $280$ videos, respectively. All five subsets contribute to the training split, while DAVIS17 and YTVOS19 additionally contribute to the test split designed for MOS-I evaluation, renamed DAVIS17-IM and YTVOS19-IM (``IM'' for Instantaneous Motion) to distinguish them from the conventional MOS datasets. Full per-subset statistics are reported in \Cref{tab:dataset-stats-full}.

\section{Experiments}
\label{sec:experiments}

\subsection{Training datasets}
\label{subsec:train_dataset}
We train \method on a composite dataset of $14{,}171$ videos ($782$K annotated frames), combining three data sources.
(\textit{i})~\textit{Synthetic data.} Kubric~\cite{Greff_2022_CVPR}, PointOdyssey~\cite{zheng2023point}, and DynamicReplica~\cite{karaev2023dynamicstereo} provide $10{,}526$ videos with pixel-perfect ground truth and motion labels.
(\textit{ii})~\textit{Real moving-object data (\method-2K).} The five \method-2K training subsets (\Cref{sec:dataset}) contribute $1{,}930$ real-world videos with temporally fine-grained motion labels.
(\textit{iii})~\textit{Real static-scene data.} $1{,}715$ clips from the Mannequin Challenge dataset~\cite{Li_2019_CVPR}, in which people remain frozen while the camera moves freely, provide static-scene supervision that teaches the model to distinguish depth parallax from independent object motion.
The detailed per-dataset statistics are provided in \Cref{tab:dataset-stats-full}.

\subsection{Benchmarks for evaluation}
\label{subsec:eval}
We evaluate \method on the following video object segmentation tasks.

\para{Moving Object Segmentation (MOS)} predicts, at every frame, a segmentation mask for every object that moves at any point in the sequence, regardless of its per-frame motion state. We evaluate on established MOS benchmarks: DAVIS16-M~\cite{Huang_2025_CVPR} (``M'' for ``Moving''), DAVIS17-M~\cite{dave2019towards}, SegTrackv2~\cite{FliICCV2013}, FBMS-59~\cite{OB14b}, and MoCA~\cite{Lamdouar20}, along with YTVOS19-M, a new benchmark derived from the \method-2K test set. Following standard practice, all benchmarks report region similarity $\mathcal{J}$ and contour accuracy $\mathcal{F}$, except MoCA, which uses detection success rate (SR)~\cite{Yang21a}.

\para{MOS-I} (``I'' for \emph{instantaneous}) is a temporally fine-grained evaluation protocol for MOS that (i) retains the sequence-level object identity association of MOS, (ii) scores segmentation masks only on frames where an object is actively moving, and (iii) explicitly penalises false-positive predictions on static or background objects. We instantiate it on two new benchmarks derived from the \method-2K test set, namely DAVIS17-IM and YTVOS19-IM (``IM'' for ``Instantaneous Motion''), and report performance on three complementary metrics.
\begin{itemize}[leftmargin=1.2em, itemindent=1pt, itemsep=1pt, parsep=2pt, topsep=-1pt]
\item \textit{Moving-object Jaccard} (${\mathcal{J}_{\text{mov}}}$) -- segmentation accuracy on actively moving frames. For each annotated object, $\mathcal{J}$ is averaged over the frames in which it is labelled as moving in \method-2K.
\item \textit{False-positive count} (FP count) -- penalisation of static and background predictions. We measure the average number of static or background objects per frame that are incorrectly predicted as moving. While $\mathcal{J}_{\text{mov}}$ rewards detecting true motion, FP count penalises spurious motion claims.
\item \textit{Mean temporal IoU} (mtIoU) -- joint measure of moving-object detection and false-positive suppression. For each sequence, we compute a temporal IoU between predicted and ground-truth moving objects, counted \textit{per object per frame}. The \emph{union} covers all (object, frame) entries marked as moving by either the prediction or the ground truth. The \emph{intersection} covers entries where a predicted and a ground-truth moving object are matched by region similarity $\mathcal{J}$ at threshold $k$. Sweeping $k \in \{0.5, 0.55, \dots, 0.95\}$ and averaging the resulting tIoUs yields the final mtIoU.
\end{itemize}

\noindentpara{Unsupervised Video Object Segmentation (UVOS)} requires automatically segmenting salient objects in a video, where targets are defined by perceptual salience and need not be moving.
We report results on DAVIS16~\cite{Perazzi16} and DAVIS17~\cite{Ponttuset17} using the standard region similarity~$\mathcal{J}$ and contour accuracy~$\mathcal{F}$.

\subsection{Implementation details}
\label{subsec:impl}

We detail the design choices of \method below. Further implementation details are provided in Appendix~\ref{appsec:imple}, alongside a comprehensive ablation study in Appendix~\ref{appsec:abla}.

\noindentpara{Architecture.}
\method uses pretrained $\pi^3$~\cite{wang2026pi} (DINOv2~\cite{oquab2024dinov} ViT-L/14) and SAM2~\cite{ravi2024sam2} (Hiera-L~\cite{ryali2023hiera}) as the geometric and segmentation encoders, both kept frozen. Bottleneck features have spatial resolution $64{\times}64$ and channel dimension $256$. The trainable transformer encoder and decoder each have $3$ layers and $8$ attention heads. We use $100$ object queries and predict masks at $256{\times}256$.

\para{Input setting.}
The \method proposer operates on $5$ frames sampled uniformly from a temporal window centred on the target frame. The window size is randomly sampled between $0.16$\,s and $1.00$\,s during training to expose the model to a range of motion speeds, and fixed at $0.5$\,s at inference.

\para{Training setup.}
We optimise with AdamW~\cite{loshchilov2018decoupled} using learning rate $1{\times}10^{-4}$, weight decay $0.01$, gradient clipping at norm $1.0$, and a cosine schedule with $2{,}000$ warmup steps. Training runs on $2$ NVIDIA RTX A6000 GPUs for approximately $2$ days, with an effective batch size of $8$. For the proposer loss in~\cref{eq:loss}, we set {\small $(\lambda_\text{mask},\,\lambda_\text{motion},\,\lambda_\text{iou},\,\lambda^{\text{mov}}_\text{conf},\,\lambda^{\text{sta}}_\text{conf}) = (20,\:1,\:1,\:0.1,\:0.001)$}, with a rescaled confidence upper bound {\small $c_{\text{max}} = 5$}. The mask loss combines focal ($\alpha{=}0.25$, $\gamma{=}2$) and dice losses in a $20{:}1$ ratio.

\begin{table*}[t]
\centering
\small
\caption{\textbf{Comparison with Moving Object Segmentation (MOS) baselines.} The{\small\setlength{\fboxsep}{1.5pt} \colorbox{gray!10}{$\Delta$ vs. prev. SoTA}} row reports the difference between our best result (\method or \method-S) and the previous state of the art.}
\vspace{-1.5mm}
\label{tab:mos}
\setlength{\tabcolsep}{4pt}
\resizebox{\linewidth}{!}{%
\begin{tabular}{lccccccccccccccc}
\toprule
\multirow{3}[3]{*}{Methods} & \multicolumn{4}{c}{Multi-Object} & \multicolumn{11}{c}{Foreground--Background} \\
\cmidrule(lr){2-5} \cmidrule(lr){6-16}
 & \multicolumn{2}{c}{YTVOS19-M} & \multicolumn{2}{c}{DAVIS17-M} & \multicolumn{2}{c}{YTVOS19-M} & \multicolumn{2}{c}{DAVIS17-M} & \multicolumn{2}{c}{DAVIS16-M} & \multicolumn{2}{c}{STv2} & \multicolumn{2}{c}{FBMS} & MoCA \\
 \cmidrule(lr){2-3} \cmidrule(lr){4-5} \cmidrule(lr){6-7} \cmidrule(lr){8-9} \cmidrule(lr){10-11} \cmidrule(lr){12-13} \cmidrule(lr){14-15} \cmidrule(lr){16-16}
 & $\mathcal{J}\uparrow$ & $\mathcal{F}\uparrow$ & $\mathcal{J}\uparrow$ & $\mathcal{F}\uparrow$ & $\mathcal{J}\uparrow$ & $\mathcal{F}\uparrow$ & $\mathcal{J}\uparrow$ & $\mathcal{F}\uparrow$ & $\mathcal{J}\uparrow$ & $\mathcal{F}\uparrow$ & $\mathcal{J}\uparrow$ & $\mathcal{F}\uparrow$ & $\mathcal{J}\uparrow$ & $\mathcal{F}\uparrow$ & SR$\uparrow$ \\
\midrule
OCLR-TTA~\cite{Xie22}  & -    & -    & $48.4$ & $49.9$ & -    & -    & $76.0$ & $75.3$ & $80.1$ & $76.9$ & $72.1$ & $75.3$ & $69.9$ & $68.3$ & $0.559$ \\
ABR~\cite{Xie24}       & -    & -    & $50.9$ & $51.2$ & -    & -    & $74.6$ & $75.2$ & $70.2$ & $73.7$ & $76.3$ & $81.1$ & $81.9$ & $79.6$ & -       \\
FlowSAM~\cite{Xie24a}   & \secondbestab{66.6} & \secondbestab{68.2} & $74.6$ & $81.0$ & $52.4$ & $54.3$ & $87.9$ & $87.0$ & $85.7$ & $83.8$ & $77.9$ & $83.9$ & $82.8$ & $80.9$ & $0.643$ \\
SegAnyMo~\cite{Huang_2025_CVPR}  & $64.6$ & $66.2$ & \secondbestab{77.4} & \secondbestab{83.6} & $70.9$ & $73.0$ & \secondbestab{90.0} & $89.0$ & \secondbestab{89.2} & $89.7$ & $76.3$ & $85.4$ & $78.3$ & $82.8$ & $0.675$ \\
GeoMotion~\cite{he2026geomotion} & -    & -    & -    & -    & $65.4$ & $65.7$ & $82.2$ & $82.3$ & $83.5$ & $84.3$ & $77.3$ & $84.3$ & $72.5$ & $78.5$ & $0.615$ \\
\midrule 
\method & \bestab{79.9} & \bestab{82.2} & \bestab{79.7} & \bestab{86.0} & \bestab{85.1} & \bestab{84.6}  & \secondbestab{90.0} & \secondbestab{90.6} & $88.7$ & \secondbestab{90.3} & \bestab{82.0} & \bestab{88.3} & \bestab{91.9} & \bestab{93.0} & \bestab{0.837} \\
\method-S & - & - & - & - & \secondbestab{81.4} & \secondbestab{81.8} & \bestab{90.7} & \bestab{93.1} & \bestab{90.0} & \bestab{93.2} & \secondbestab{78.9} & \secondbestab{85.9} & \secondbestab{83.0} & \secondbestab{84.9} & \secondbestab{0.785} \\ 
\midrule
{\samevalue{$\Delta$ vs. prev. SoTA}}
 & {\footnotesize \improv{$+13.3$}} & {\footnotesize \improv{$+14.0$}}
 & {\footnotesize \improv{$+2.3$}}  & {\footnotesize \improv{$+2.4$}}
 & {\footnotesize \improv{$+14.2$}} & {\footnotesize \improv{$+11.6$}}
 & {\footnotesize \improv{$+0.7$}}  & {\footnotesize \improv{$+4.1$}}
 & {\footnotesize \improv{$+0.8$}}  & {\footnotesize \improv{$+3.5$}}
 & {\footnotesize \improv{$+4.1$}}  & {\footnotesize \improv{$+2.9$}}
 & {\footnotesize \improv{$+9.1$}}  & {\footnotesize \improv{$+10.2$}}
 & {\footnotesize \improv{$+0.162$}} \\
\bottomrule
\end{tabular}%
}
\vspace{-1.5mm}
\end{table*}

\subsection{Quantitative results}

\Cref{tab:mos} reports results on the conventional MOS setting, where moving-object masks are predicted at the sequence level. \method, and its foreground--background variant \method-S, achieves new state-of-the-art performance across all benchmarks. Gains are most pronounced on YTVOS19-M, whose denser and more varied object interactions expose the limitations of prior flow- and trajectory-based methods. \method also remains robust on camouflaged objects (MoCA), where segmentation must rely exclusively on motion cues.

\begin{table*}[t]
\centering
\small
\begin{minipage}{0.573\linewidth}
\centering
\caption{\textbf{Comparison with MOS-I baselines.} ``I'' is short for ``instantaneous''.}
\vspace{-1.5mm}
\label{tab:tmos}
\setlength{\tabcolsep}{4pt}
\resizebox{\linewidth}{!}{%
\begin{tabular}{lcccccc}
\toprule
\multirow{2}[4]{*}{Methods} & \multicolumn{3}{c}{YTVOS19-IM} & \multicolumn{3}{c}{DAVIS17-IM} \\
\cmidrule(lr){2-4} \cmidrule(lr){5-7}
 & mtIoU$\uparrow$ & $\mathcal{J}_{\text{mov}}\!\uparrow$ & \makecell{FP\\count}$\downarrow$ & mtIoU$\uparrow$ & $\mathcal{J}_{\text{mov}}\!\uparrow$ & \makecell{FP\\count}$\downarrow$ \\
\midrule
OCLR-TTA~\cite{Xie22} & -      & -      & -       & $28.3$ & $54.1$ & $0.550$ \\
ABR~\cite{Xie24}      & -      & -      & -       & $36.0$ & $57.0$ & $0.271$ \\
FlowSAM~\cite{Xie24a}  & $21.1$ & $62.8$ & $3.085$ & $22.7$ & $75.3$ & $4.101$ \\
SegAnyMo~\cite{Huang_2025_CVPR} & $55.9$ & $63.2$ & $0.320$ & $67.0$ & $81.7$ & $0.253$ \\
\midrule
\method (\textit{online}) & \bestab{64.2} & \secondbestab{76.4} & \bestab{0.197}  & \bestab{72.9} & \secondbestab{83.5} & \bestab{0.075} \\
\method                 & \secondbestab{63.4} & \bestab{76.6} & \secondbestab{0.238} & \bestab{72.9} & \bestab{84.2} & \secondbestab{0.079} \\
\midrule
{\samevalue{$\Delta$ vs. prev. SoTA}}
 & {\footnotesize \improv{$+8.3$}} & {\footnotesize \improv{$+13.4$}} & {\footnotesize \improv{$+0.123$}}
 & {\footnotesize \improv{$+5.9$}} & {\footnotesize \improv{$+2.5$}}  & {\footnotesize \improv{$+0.178$}} \\
\bottomrule
\end{tabular}}
\end{minipage}
\hfill
\begin{minipage}{0.395\linewidth}
\centering
\caption{\textbf{Comparison with Unsupervised VOS (UVOS) baselines.}}
\vspace{-1.5mm}
\label{tab:uvos}
\setlength{\tabcolsep}{4.7pt}
\resizebox{\linewidth}{!}{%
\begin{tabular}{lcccc}
\toprule
\multirow{2}[2]{*}{Methods} & \multicolumn{2}{c}{DAVIS17} & \multicolumn{2}{c}{DAVIS16} \\
\cmidrule(lr){2-3} \cmidrule(lr){4-5}
 & \;\;$\mathcal{J}\uparrow$\;\; & \;\;$\mathcal{F}\uparrow$\;\; & \;\;$\mathcal{J}\uparrow$\;\; & \;\;$\mathcal{F}\uparrow$\;\; \\
\midrule  
OCLR-TTA~\cite{Xie22}  & $43.7$  & $44.5$ & $80.8$ & $76.8$ \\
ABR~\cite{Xie24}       & $45.0$  & $45.3$ & $71.8$ & $73.2$ \\
DEVA~\cite{cheng2023tracking}      & $70.4$ & \secondbestab{76.4} & $87.6$ & $90.2$ \\
FlowSAM~\cite{Xie24a}   & \secondbestab{71.2} & $76.3$ & $87.1$ & $84.9$ \\
SegAnyMo~\cite{Huang_2025_CVPR}  & $66.5$ & $71.4$ & \bestab{90.6} & \secondbestab{91.0} \\
GeoMotion~\cite{he2026geomotion} & -    & -    & $84.5$ & $85.0$ \\
\midrule
\method & \bestab{76.2} & \bestab{80.6} & \secondbestab{90.1} & \bestab{91.5} \\
\method-S & -    & - & $87.9$ & $90.8$\\
\midrule
{\samevalue{$\Delta$ vs. prev. SoTA}}
 & {\footnotesize \improv{$+5.0$}} & {\footnotesize \improv{$+4.2$}}
 & {\footnotesize \regress{$-0.5$}} & {\footnotesize \improv{$+0.5$}} \\
\bottomrule
\end{tabular}}
\end{minipage}
\vspace{-1mm}
\end{table*}

As shown in \Cref{tab:tmos}, performance gaps widen further on MOS-I, which requires predictions at finer temporal granularity. Since prior methods treat motion as a sequence-level attribute and cannot capture instantaneous motion states, \method surpasses them by a substantial margin on all three metrics.
We additionally evaluate an online variant of \method (\Cref{subsec:propagator}), which performs on par with the default offline mode. This can be attributed to the instantaneous nature of MOS-I, where a short causal window already provides sufficient motion evidence. The offline mode's second global propagation even slightly hurts FP count and mtIoU, as the bidirectional pass amplifies any borderline false positives from the first pass.
The online mode, however, is restricted to the MOS-I setup, since a single forward pass cannot recover masks for objects that are initially static and only later begin to move. The offline procedure therefore remains our default across all tasks, including MOS and UVOS.

We also report performance on Unsupervised Video Object Segmentation (UVOS), which targets the prediction of salient objects regardless of their motion state. Although UVOS lies outside our primary focus of segmenting only moving objects, \Cref{tab:uvos} shows that \method generalises well to this adjacent task, achieving state-of-the-art results on the multi-object DAVIS17 benchmark and remaining on par with the prior best on the single-object DAVIS16.

\subsection{Qualitative results}
\begin{figure*}[t]
\centering
\includegraphics[width=0.98\linewidth, trim=0 0 0 0, clip]{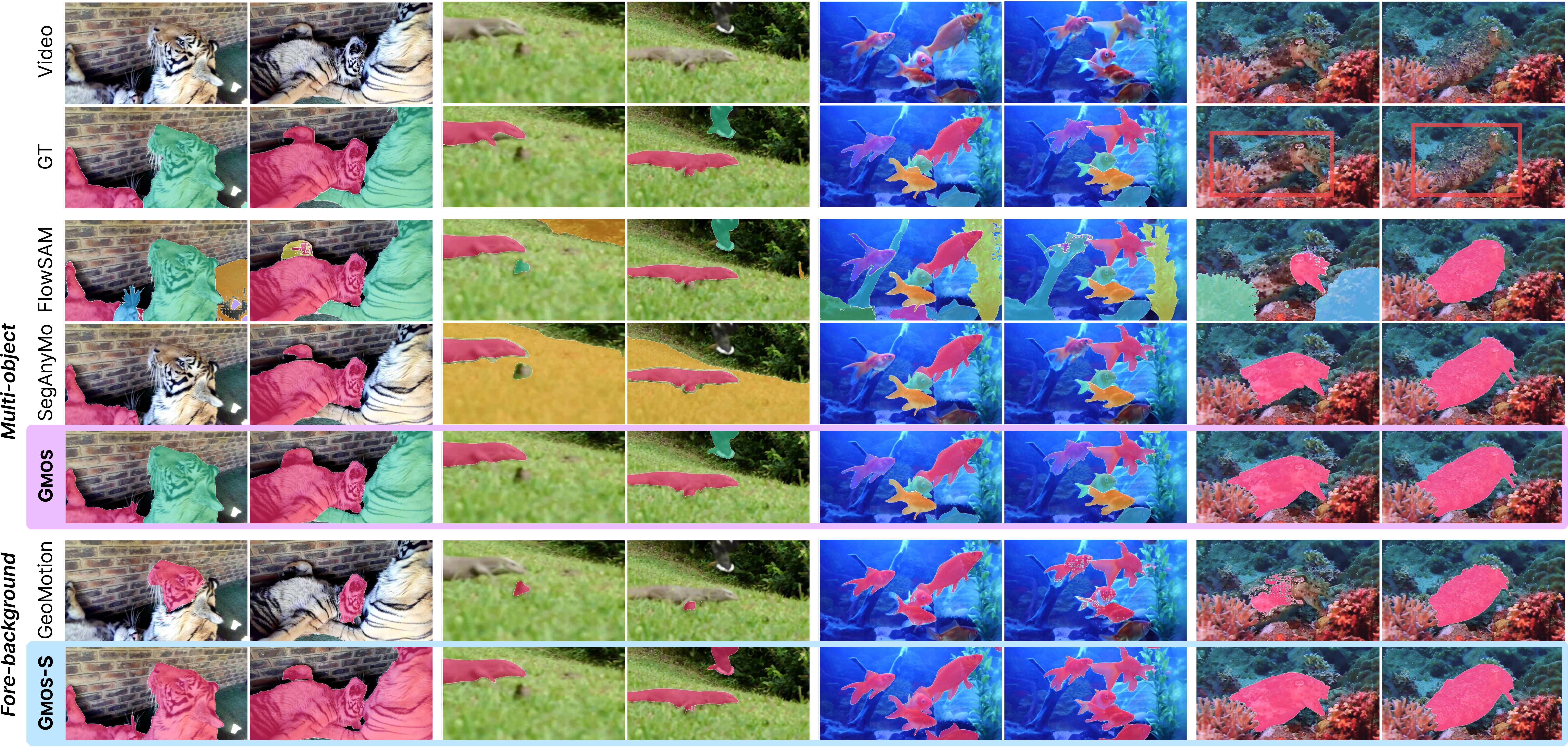}
\vspace{-1mm}
\caption{
\textbf{Qualitative comparison on the MOS task.}
Example videos are sampled from YTVOS19 (first two columns), DAVIS17 (third column), and MoCA (last column). The middle block shows multi-object results, and the bottom block shows foreground--background results. }

\label{fig:qual}
\vspace{-3mm}
\end{figure*}

\Cref{fig:qual} compares \method against representative baselines on sequences from YTVOS19, DAVIS17, and MoCA. Regarding multi-object predictions, FlowSAM~\cite{Xie24a} tends to over-segment moving objects (\eg treating the coral as moving), while SegAnyMo~\cite{Huang_2025_CVPR} misses moving objects or bleeds masks into the background. In contrast, \method assigns each moving object a consistent identity and boundary that closely follow the ground truth, even under heavy occlusion. Under the foreground--background protocol, GeoMotion~\cite{he2026geomotion} captures prominent motion but overlooks instances near image boundaries (\eg the right tiger), whereas \method-S recovers them reliably.

\begin{figure*}[t]
\includegraphics[width=1.0\linewidth, trim=0 0 0 0, clip]{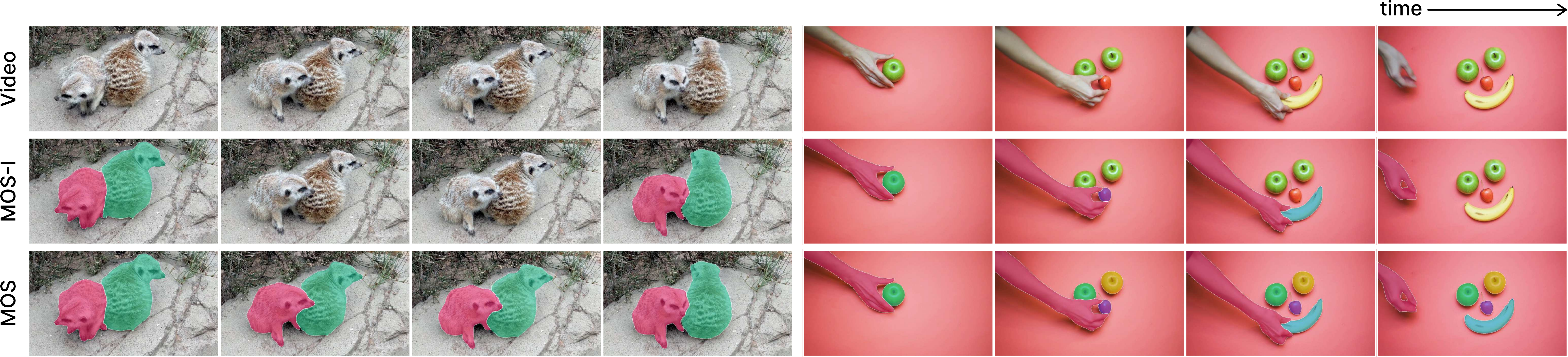}
\vspace{-4.5mm}
\caption{
\textbf{\method on in-the-wild videos.} Two in-the-wild sequences (sourced outside our training or test datasets) illustrate our MOS and MOS-I predictions. Under MOS-I, instantaneously moving objects are segmented, with object identities consistently associated across frames. Under MOS, \method produces full segmentation masks for every object that moves at any point in the sequence, regardless of its motion state at a given frame. 
}
\vspace{-2mm}
\label{fig:qual_main_wild}
\end{figure*}

\Cref{fig:qual_main_wild} illustrates \method on in-the-wild videos, contrasting MOS-I and MOS predictions. In the left sequence, the meerkats become static mid-sequence. Our MOS-I predictions correctly identify the static states, while MOS produces consistent segmentation across frames. In the right sequence, a hand arranges fruits one piece at a time. Our MOS-I predictions correctly release each piece once it comes to rest, whereas MOS keeps them masked throughout.

\section{Related works}
\label{sec:related_works}

\noindentpara{Unsupervised Video Object Segmentation (UVOS)}
aims to identify salient objects in videos without any annotation at inference time, in contrast to semi-supervised VOS~\cite{Perazzi16, Ponttuset17, Xu18, iccv19_stm, cheng2022xmem, yang2022deaot, Cheng_2024_CVPR, ravi2024sam2}, which propagates masks from first-frame ground truths.
Early methods adopt two-stream architectures that fuse motion and appearance inputs~\cite{Jain17, Lu_2019_CVPR, zhou20, amcnet, Ren_2021_CVPR, Liu_Yu_Wang_Zhou_2021, Ji_2021_ICCV, transportnet, HFAN}, while others rely primarily on saliency or attention mechanisms~\cite{Wang19, Yang19, Fan19, pmn, tmo, Yuan_2023_ICCV, Zhou_2021_CVPR, Liu_2024_CVPR, Cho_2024_CVPR, Lee_2024_CVPR, Song_Su_Zheng_Zhang_Liu_Liu_2024}.
More recent works scale UVOS with new data strategies and modular pipeline designs: VideoCutLER~\cite{wang2023videocutler} scales training via copy-paste augmentation transferred from unsupervised image instance segmentation, DEVA~\cite{cheng2023tracking} composes an image-level segmenter with a separate temporal propagator, and SAM-based approaches~\cite{kirillov2023segment, ravi2024sam2, carion2026sam, zhang2024uvosam} build UVOS pipelines around pre-trained foundation segmenters.
Despite significant progress, UVOS methods typically target the most {salient} objects, which do not necessarily correspond to objects that are actually moving.

\vspace{1mm}
\para{Motion Segmentation (MS)}
seeks to separate independently moving regions from the static background, producing a binary foreground--background mask.
Mainstream approaches take optical flow as the motion signal and cluster pixels according to similar motion patterns~\cite{Papazoglou13, Bideau16, Tokmakov17, Tokmakov19, Yang21a, Lamdouar20, dave2019towards, Yang_2021_CVPR, Choudhury22, yang_loquercio_2019, meunier2022driven, meunier2023bunsupervised}.
However, these flow-based approaches are inherently bound by the quality of the pre-computed optical flow, and also vulnerable to large camera motion and induced depth parallax.
A complementary line of work aggregates motion cues over longer temporal windows via point trajectories~\cite{Brox10, Ochs11, Keuper15, Fragkiadaki12}. Recent advances in point tracking~\cite{Karaev25, tapvid, Doersch_2023_ICCV} have substantially improved long-range correspondence, enabling segmentation to be learnt directly from dense trajectories~\cite{Karazija24b, Huang_2025_CVPR}. However, such reliance on long-term tracking makes these methods insensitive to static--dynamic transitions, limiting their online applicability.

Another line of work exploits 3D or 4D geometric priors for motion segmentation. MegaSAM~\cite{Li_2025_CVPR} learns motion probability maps from estimated camera parameters and depth to separate moving and static content, while RoMo~\cite{golisabour2025romo} combines optical flow with epipolar constraints and iteratively refines motion masks using segmentation priors from SAM~\cite{kirillov2023segment}. Building upon feed-forward reconstruction methods~\cite{Wang_2024_CVPR, Wang25}, MonST3R~\cite{zhang2025monstr} and DAS3R~\cite{xu2024das3r} jointly decode moving regions alongside temporally consistent video reconstructions. Easi3R~\cite{chen2025easi3r} reveals that the attention layers of such reconstruction models already encode relative camera-object motion. GeoMotion~\cite{he2026geomotion} further exploits the latent 4D features of $\pi^3$~\cite{wang2026pi} to directly disentangle object and camera motion in a feed-forward manner.

\vspace{1mm}
\para{Moving Object Segmentation (MOS)}
aims to discover, segment, and track individual moving objects. For videos containing a single moving object, MOS degenerates to foreground--background motion segmentation. Recent works instead focus on the more challenging multi-object scenario, similarly building upon the aforementioned motion cues (\ie flow, trajectories, and geometric priors).

Without relying on human annotations, one group of methods~\cite{Safadoust23, Lian_2023_CVPR, lao2025dividedattentionunsupervisedmultiobject} pursues purely unsupervised training via flow reconstruction as a proxy objective. Alternatively, OCLR~\cite{Xie22} takes optical flow as input and trains object-centric layered representations on synthetic data to discover and track multiple objects under occlusion, while ABR~\cite{Xie24} extends it with appearance-based refinement.

With increasingly available annotated masks and strong pre-trained models, FlowSAM~\cite{Xie24a} injects optical flow into SAM~\cite{kirillov2023segment}, either as an input channel or as prompt signals, for moving object segmentation. SegAnyMo~\cite{Huang_2025_CVPR} fuses long-range trajectory cues with DINO~\cite{caron2021emerging} features and employs SAM2~\cite{ravi2024sam2} for mask densification.

\section{Conclusion}
\label{sec:conclusion}

We presented \method, a feed-forward framework that grounds Moving Object Segmentation in 3D space and time, making it robust to camera-induced parallax in space and temporally sensitive to per-frame motion-state changes. \method operates on RGB video alone and supports both online and offline inference. To facilitate training and evaluation in this regime, we curated \method-2K, a real-world video dataset with per-object temporal motion annotations, and formalised MOS-I, a temporally fine-grained evaluation protocol with three complementary metrics. \method achieves state-of-the-art results on MOS, MOS-I, and UVOS benchmarks, while running roughly three times faster than prior MOS methods.

\para{Acknowledgements.} This research is supported by the UK EPSRC Programme Grant VisualAI (EP/T028572/1), the Royal Society Research Professorship RSRP$\backslash$R$\backslash$241003, and the Clarendon Scholarship.

\bibliographystyle{plainnat}
\bibliography{references, vgg_local}

@String(PAMI = {IEEE Trans. Pattern Anal. Mach. Intell.})

@String(IJCV = {Int. J. Comput. Vis.})

@String(CVPR= {IEEE Conf. Comput. Vis. Pattern Recog.})

@String(ICCV= {Int. Conf. Comput. Vis.})

@String(ECCV= {Eur. Conf. Comput. Vis.})

@String(NIPS= {Adv. Neural Inform. Process. Syst.})

@String(BMVC= {Brit. Mach. Vis. Conf.})

@String(ACMMM= {ACM Int. Conf. Multimedia})

@String(ACCV  = {ACCV})

@String(ICLR = {Int. Conf. Learn. Represent.})

@String(AAAI = {AAAI})

@String(PAMI  = {IEEE TPAMI})

@String(IJCV  = {IJCV})

@String(CVPR  = {CVPR})

@String(ICCV  = {ICCV})

@String(ECCV  = {ECCV})

@String(NIPS  = {NeurIPS})

@String(NEURIPS  = {NeurIPS})

@String(BMVC  =	{BMVC})

@String(ACMMM = {ACM MM})

@String(ICLR  = {ICLR})

@String(WACV= {WACV})

@inproceedings{Perazzi16,
  author={Perazzi, Federico and Pont-Tuset, Jordi and McWilliams, Brian and Van Gool, Luc and Gross, Markus and Sorkine-Hornung, Alexander},
  title={A benchmark dataset and evaluation methodology for video object segmentation},
  booktitle =   "CVPR",
  year      =   "2016",
}

@article{Ponttuset17,
  author    =  "Jordi Pont-Tuset and Federico Perazzi and Sergi Caelles and Pablo Arbeláez and Alex Sorkine-Hornung and Luc Van Gool",
  title     =   "The 2017 davis challenge on video object segmentation",
  journal   =   "arXiv preprint arXiv:1704.00675",
  year      =   "2017",
}

@inproceedings{Xu18,
  title={Youtube-vos: A large-scale video object segmentation benchmark},
  author={Xu, Ning and Yang, Linjie and Fan, Yuchen and Yue, Dingcheng and Liang, Yuchen and Yang, Jianchao and Huang, Thomas},
  booktitle={ECCV},
  year={2018}
}

@inproceedings{iccv19_stm,
  title={Video object segmentation using space-time memory networks},
  author={Oh, Seoung Wug and Lee, Joon-Young and Xu, Ning and Kim, Seon Joo},
  booktitle={ICCV},
  year={2019}
}

@inproceedings{cheng2022xmem,
  title={{XMem}: Long-Term Video Object Segmentation with an Atkinson-Shiffrin Memory Model},
  author={Cheng, Ho Kei and Alexander G. Schwing},
  booktitle=ECCV,
  year={2022}
}

@inproceedings{yang2022deaot,
  title={Decoupling Features in Hierarchical Propagation for Video Object Segmentation},
  author={Yang, Zongxin and Yang, Yi},
  booktitle=NIPS,
  year={2022}
}

@article{ravi2024sam2,
  title={SAM 2: Segment Anything in Images and Videos},
  author={Ravi, Nikhila and Gabeur, Valentin and Hu, Yuan-Ting and Hu, Ronghang and Ryali, Chaitanya and Ma, Tengyu and Khedr, Haitham and R{\"a}dle, Roman and Rolland, Chloe and Gustafson, Laura and Mintun, Eric and Pan, Junting and Alwala, Kalyan Vasudev and Carion, Nicolas and Wu, Chao-Yuan and Girshick, Ross and Doll{\'a}r, Piotr and Feichtenhofer, Christoph},
  journal={arXiv preprint arXiv:2408.00714},
  year={2024}
}

@inproceedings{Jain17,
  title={Fusionseg: Learning to combine motion and appearance for fully automatic segmentation of generic objects in videos},
  author={Jain, Suyog Dutt and Xiong, Bo and Grauman, Kristen},
  booktitle =   {CVPR},
  year      =   "2017",
}

@inproceedings{Lu_2019_CVPR,  
    author = {Lu, Xiankai and Wang, Wenguan and Ma, Chao and Shen, Jianbing and Shao, Ling and Porikli, Fatih},  
    title = {See More, Know More: Unsupervised Video Object Segmentation With Co-Attention Siamese Networks},  
    booktitle = {CVPR},  
    year = {2019}  
}

@inproceedings{zhou20,
  title={Motion-attentive transition for zero-shot video object segmentation},
  author={Tianfei Zhou and Shunzhou Wang and Yi Zhou and Yazhou Yao and Jianwu Li and Ling Shao},
  booktitle=AAAI,
  year={2020}
}

@INPROCEEDINGS {amcnet,
author = {S. Yang and L. Zhang and J. Qi and H. Lu and S. Wang and X. Zhang},
booktitle = ICCV,
title = {Learning Motion-Appearance Co-Attention for Zero-Shot Video Object Segmentation},
year = {2021}
}

@inproceedings{transportnet,
author = {K. Zhang and Z. Zhao and D. Liu and Q. Liu and B. Liu},
booktitle = ICCV,
title = {Deep Transport Network for Unsupervised Video Object Segmentation},
year = {2021}
}

@inproceedings{Yang19,
    title={Anchor Diffusion for Unsupervised Video Object Segmentation},
    author={Zhao Yang and Qiang Wang and Luca Bertinetto and Song Bai and Weiming Hu and Philip H.S. Torr},
    booktitle={ICCV},
    year={2019}
}

@inproceedings{Fan19,
  title={Shifting more attention to video salient object detection},
  author={Fan, Deng-Ping and Wang, Wenguan and Cheng, Ming-Ming and Shen, Jianbing},
  booktitle={CVPR},
  year={2019}
}

@inproceedings{pmn,
    author = {M. Lee and S. Cho and S. Lee and C. Park and S. Lee},
    booktitle = WACV,
    title = {Unsupervised Video Object Segmentation via Prototype Memory Network},
    year = {2023}
}

@inproceedings{tmo,
    author = {S. Cho and M. Lee and S. Lee and C. Park and D. Kim and S. Lee},
    booktitle = WACV,
    title = {Treating Motion as Option to Reduce Motion Dependency in Unsupervised Video Object Segmentation},
    year = {2023},
}

@article{wang2023videocutler,
  title={VideoCutLER: Surprisingly Simple Unsupervised Video Instance Segmentation},
  author={Wang, Xudong and Misra, Ishan and Zeng, Zizun and Girdhar, Rohit and Darrell, Trevor},
  journal={arXiv preprint arXiv:2308.14710},
  year={2023}
}

@inproceedings{cheng2023tracking,
  title={Tracking Anything with Decoupled Video Segmentation},
  author={Cheng, Ho Kei and Oh, Seoung Wug and Price, Brian and Schwing, Alexander and Lee, Joon-Young},
  booktitle=ICCV,
  year={2023}
}

@inproceedings{kirillov2023segment,
    author    = {Kirillov, Alexander and Mintun, Eric and Ravi, Nikhila and Mao, Hanzi and Rolland, Chloe and Gustafson, Laura and Xiao, Tete and Whitehead, Spencer and Berg, Alexander C. and Lo, Wan-Yen and Dollar, Piotr and Girshick, Ross},
    title     = {Segment Anything},
    booktitle = ICCV,
    year      = {2023}
}

@article{zhang2024uvosam,
    title={UVOSAM: A Mask-free Paradigm for Unsupervised Video Object Segmentation via Segment Anything Model}, 
    author={Zhenghao Zhang and Shengfan Zhang and Zhichao Wei and Zuozhuo Dai and Siyu Zhu},
    year={2024},
    journal={arXiv preprint arXiv:2305.12659},
}

@inproceedings{Cheng_2024_CVPR,
    author    = {Cheng, Ho Kei and Oh, Seoung Wug and Price, Brian and Lee, Joon-Young and Schwing, Alexander},
    title     = {Putting the Object Back into Video Object Segmentation},
    booktitle = CVPR,
    year      = {2024},
}

@inproceedings{carion2026sam,
    title={{SAM} 3: Segment Anything with Concepts},
    author={Nicolas Carion and Laura Gustafson and Yuan-Ting Hu and Shoubhik Debnath and Ronghang Hu and Didac Suris Coll-Vinent and Chaitanya Ryali and Kalyan Vasudev Alwala and Haitham Khedr and Andrew Huang and Jie Lei and Tengyu Ma and Baishan Guo and Arpit Kalla and Markus Marks and Joseph Greer and Meng Wang and Peize Sun and Roman R{\"a}dle and Triantafyllos Afouras and Effrosyni Mavroudi and Katherine Xu and Tsung-Han Wu and Yu Zhou and Liliane Momeni and RISHI HAZRA and Shuangrui Ding and Sagar Vaze and Francois Porcher and Feng Li and Siyuan Li and Aishwarya Kamath and Ho Kei Cheng and Piotr Dollar and Nikhila Ravi and Kate Saenko and Pengchuan Zhang and Christoph Feichtenhofer},
    booktitle=ICLR,
    year={2026},
}

@inproceedings{Ren_2021_CVPR,
    author    = {Ren, Sucheng and Liu, Wenxi and Liu, Yongtuo and Chen, Haoxin and Han, Guoqiang and He, Shengfeng},
    title     = {Reciprocal Transformations for Unsupervised Video Object Segmentation},
    booktitle = CVPR,
    year      = {2021},
}

@inproceedings{Liu_Yu_Wang_Zhou_2021, 
    title={F2Net: Learning to Focus on the Foreground for Unsupervised Video Object Segmentation}, 
    booktitle=AAAI, 
    author={Liu, Daizong and Yu, Dongdong and Wang, Changhu and Zhou, Pan}, year={2021},
}

@inproceedings{Ji_2021_ICCV,
    author    = {Ji, Ge-Peng and Fu, Keren and Wu, Zhe and Fan, Deng-Ping and Shen, Jianbing and Shao, Ling},
    title     = {Full-Duplex Strategy for Video Object Segmentation},
    booktitle = ICCV,
    year      = {2021},
}

@inproceedings{HFAN,
    author="Pei, Gensheng
    and Shen, Fumin
    and Yao, Yazhou
    and Xie, Guo-Sen
    and Tang, Zhenmin
    and Tang, Jinhui",
    title="Hierarchical Feature Alignment Network for Unsupervised Video Object Segmentation",
    booktitle= ECCV,
    year="2022",
}

@inproceedings{Yuan_2023_ICCV,
    author    = {Yuan, Yichen and Wang, Yifan and Wang, Lijun and Zhao, Xiaoqi and Lu, Huchuan and Wang, Yu and Su, Weibo and Zhang, Lei},
    title     = {Isomer: Isomerous Transformer for Zero-shot Video Object Segmentation},
    booktitle = ICCV,
    year      = {2023},
}

@inproceedings{Zhou_2021_CVPR,
    author    = {Zhou, Tianfei and Li, Jianwu and Li, Xueyi and Shao, Ling},
    title     = {Target-Aware Object Discovery and Association for Unsupervised Video Multi-Object Segmentation},
    booktitle = CVPR,
    year      = {2021},
}

@inproceedings{Liu_2024_CVPR,
    author    = {Liu, Weihuang and Shen, Xi and Li, Haolun and Bi, Xiuli and Liu, Bo and Pun, Chi-Man and Cun, Xiaodong},
    title     = {Depth-aware Test-Time Training for Zero-shot Video Object Segmentation},
    booktitle = CVPR,
    year      = {2024},
}

@inproceedings{Cho_2024_CVPR,
    author    = {Cho, Suhwan and Lee, Minhyeok and Lee, Seunghoon and Lee, Dogyoon and Choi, Heeseung and Kim, Ig-Jae and Lee, Sangyoun},
    title     = {Dual Prototype Attention for Unsupervised Video Object Segmentation},
    booktitle = CVPR,
    year      = {2024},
}

@inproceedings{Lee_2024_CVPR,
    author    = {Lee, Minhyeok and Cho, Suhwan and Lee, Dogyoon and Park, Chaewon and Lee, Jungho and Lee, Sangyoun},
    title     = {Guided Slot Attention for Unsupervised Video Object Segmentation},
    booktitle = CVPR,
    year      = {2024},
}

@inproceedings{Song_Su_Zheng_Zhang_Liu_Liu_2024, 
    title={Generalizable Fourier Augmentation for Unsupervised Video Object Segmentation},  
    booktitle=AAAI, 
    author={Song, Huihui and Su, Tiankang and Zheng, Yuhui and Zhang, Kaihua and Liu, Bo and Liu, Dong}, 
    year={2024}, 
}

@inproceedings{Papazoglou13,
  title={Fast object segmentation in unconstrained video},
  author={Anestis Papazoglou and Vittorio Ferrari},
  booktitle={ICCV},
  year={2013}
}

@inproceedings{Bideau16,
  author    =   {Bideau, Pia and Learned-Miller, Erik},
  title     =   {It’s moving! A probabilistic model for causal motion segmentation in moving camera videos},
  booktitle =   {ECCV},
  year      =   "2016",
}

@inproceedings{Tokmakov17,
  author={Pavel Tokmakov and Karteek Alahari and Cordelia Schmid},
  title={Learning Video Object Segmentation with Visual Memory},
  booktitle = ICCV,
  year = {2017}
}

@article{Tokmakov19,
  author={Tokmakov, Pavel and Schmid, Cordelia and Alahari, Karteek},
  title={Learning to segment moving objects},
  journal   =   IJCV,
  year      =   "2019",
}

@inproceedings{dave2019towards,
  title={Towards segmenting anything that moves},
  author={Dave, Achal and Tokmakov, Pavel and Ramanan, Deva},
  booktitle=ICCV,
  year={2019}
}

@inproceedings{Yang_2021_CVPR,
    author    = {Yang, Yanchao and Lai, Brian and Soatto, Stefano},
    title     = {DyStaB: Unsupervised Object Segmentation via Dynamic-Static Bootstrapping},
    booktitle = CVPR,
    year      = {2021},
}

@inproceedings{yang_loquercio_2019,
  title={Unsupervised Moving Object Detection via Contextual Information Separation},
  author={Yang, Yanchao and Loquercio, Antonio and Scaramuzza, Davide and Soatto, Stefano},
  booktitle = {CVPR},
  year={2019},
}

@article{meunier2022driven,
  title={EM-driven unsupervised learning for efficient motion segmentation},
  author={Meunier, Etienne and Badoual, Ana{\"\i}s and Bouthemy, Patrick},
  journal=PAMI,
  year={2022},
}

@article{meunier2023bunsupervised,
  title={Unsupervised motion segmentation in one go: Smooth long-term model over a video},
  author={Meunier, Etienne and Bouthemy, Patrick},
  journal={arXiv preprint arXiv:2310.01040},
  year={2023}
}

@inproceedings{Brox10,
  author={Brox, Thomas and Malik, Jitendra},
  title={Object segmentation by long term analysis of point trajectories},
  booktitle =   {ECCV},
  year      =   "2010",
}

@inproceedings{Ochs11,
  author={Ochs, Peter and Brox, Thomas},
  title={Object segmentation in video: a hierarchical variational approach for turning point trajectories into dense regions},
  booktitle =   {ICCV},
  year      =   "2011",
}

@inproceedings{Keuper15,
  author    =  "Margret Keuper and Bjoern Andres and Thomas Brox",
  title     = "Motion trajectory segmentation via minimum cost multicuts",
  booktitle ={ICCV},
  year      = "2015",
}

@inproceedings{Fragkiadaki12,
  author    =  "Katerina Fragkiadaki and Geng Zhang and Jianbo Shi",
  title     = "Video segmentation by tracing discontinuities in a trajectory embedding",
  booktitle = {CVPR},
  year      = "2012",
}

@inproceedings{tapvid,
 author = {Doersch, Carl and Gupta, Ankush and Markeeva, Larisa and Recasens, Adria and Smaira, Lucas and Aytar, Yusuf and Carreira, Joao and Zisserman, Andrew and Yang, Yi},
 booktitle = NIPS,
 title = {TAP-Vid: A Benchmark for Tracking Any Point in a Video},
 year = {2022}
}

@inproceedings{Doersch_2023_ICCV,
    author    = {Doersch, Carl and Yang, Yi and Vecerik, Mel and Gokay, Dilara and Gupta, Ankush and Aytar, Yusuf and Carreira, Joao and Zisserman, Andrew},
    title     = {TAPIR: Tracking Any Point with Per-Frame Initialization and Temporal Refinement},
    booktitle = ICCV,
    year      = {2023},
}

@inproceedings{Wang_2024_CVPR,
    author    = {Wang, Shuzhe and Leroy, Vincent and Cabon, Yohann and Chidlovskii, Boris and Revaud, Jerome},
    title     = {DUSt3R: Geometric 3D Vision Made Easy},
    booktitle = CVPR,
    year      = {2024},
}

@article{xu2024das3r,
 title     = {DAS3R: Dynamics-Aware Gaussian Splatting for Static Scene Reconstruction}, 
 author    = {Xu, Kai and Tse, Tze Ho Elden and Peng, Jizong and Yao, Angela},
 journal   = {arXiv preprint arxiv:2412.19584},
 year      = {2024}
}

@inproceedings{chen2025easi3r,
    title={Easi3R: Estimating Disentangled Motion from DUSt3R Without Training},
    author={Chen, Xingyu and Chen, Yue and Xiu, Yuliang and Geiger, Andreas and Chen, Anpei},
    booktitle=ICCV,
    year={2025}
}

@inproceedings{Li_2025_CVPR,
    author    = {Li, Zhengqi and Tucker, Richard and Cole, Forrester and Wang, Qianqian and Jin, Linyi and Ye, Vickie and Kanazawa, Angjoo and Holynski, Aleksander and Snavely, Noah},
    title     = {MegaSaM: Accurate, Fast and Robust Structure and Motion from Casual Dynamic Videos},
    booktitle = CVPR,
    year      = {2025},
}

@inproceedings{golisabour2025romo,
    title={{RoMo}: Robust Motion Segmentation Improves Structure from Motion},
    author={Goli, Lily and Sabour, Sara and Matthews, Mark and Marcus, Brubaker and Lagun, Dmitry and Jacobson, Alec and Fleet, David J. and Saxena, Saurabh and Tagliasacchi, Andrea},
    booktitle=ICCV,
    year={2025}
}

@inproceedings{he2026geomotion,
      title={GeoMotion: Rethinking Motion Segmentation via Latent 4D Geometry}, 
      author={Xiankang He and Peile Lin and Ying Cui and Dongyan Guo and Chunhua Shen and Xiaoqin Zhang},
      year={2026},
      booktitle=CVPR,
}

@inproceedings{wang2026pi,
    title={$\pi^3$: Permutation-Equivariant Visual Geometry Learning},
    author={Yifan Wang and Jianjun Zhou and Haoyi Zhu and Wenzheng Chang and Yang Zhou and Zizun Li and Junyi Chen and Jiangmiao Pang and Chunhua Shen and Tong He},
    booktitle=ICLR,
    year={2026},
}

@inproceedings{zhang2025monstr,
    title={Mon{ST}3R: A Simple Approach for Estimating Geometry in the Presence of Motion},
    author={Junyi Zhang and Charles Herrmann and Junhwa Hur and Varun Jampani and Trevor Darrell and Forrester Cole and Deqing Sun and Ming-Hsuan Yang},
    booktitle=ICLR,
    year={2025},
}

@inproceedings{Safadoust23,
    title={Multi-Object Discovery by Low-Dimensional Object Motion},
    author={Sadra Safadoust and Fatma Güney},
    booktitle=ICCV,
    year={2023}
}

@inproceedings{caron2021emerging,
  title={Emerging Properties in Self-Supervised Vision Transformers},
  author={Caron, Mathilde and Touvron, Hugo and Misra, Ishan and J\'egou, Herv\'e  and Mairal, Julien and Bojanowski, Piotr and Joulin, Armand},
  booktitle=ICCV,
  year={2021}
}

@inproceedings{Lian_2023_CVPR,
    author    = {Lian, Long and Wu, Zhirong and Yu, Stella X.},
    title     = {Bootstrapping Objectness From Videos by Relaxed Common Fate and Visual Grouping},
    booktitle = CVPR,
    year      = {2023},
}

@article{lao2025dividedattentionunsupervisedmultiobject,
    title={Divided Attention: Unsupervised Multi-Object Discovery with Contextually Separated Slots}, 
    author={Dong Lao and Zhengyang Hu and Francesco Locatello and Yanchao Yang and Stefano Soatto},
    year={2023},
    journal = {arXiv preprint arxiv:2304.01430},
}

@inproceedings{Huang_2025_CVPR,
    author    = {Huang, Nan and Zheng, Wenzhao and Xu, Chenfeng and Keutzer, Kurt and Zhang, Shanghang and Kanazawa, Angjoo and Wang, Qianqian},
    title     = {Segment Any Motion in Videos},
    booktitle = CVPR,
    year      = {2025},
}

@inproceedings{Greff_2022_CVPR,
    author    = {Greff, Klaus and Belletti, Francois and Beyer, Lucas and Doersch, Carl and Du, Yilun and Duckworth, Daniel and Fleet, David J. and Gnanapragasam, Dan and Golemo, Florian and Herrmann, Charles and Kipf, Thomas and Kundu, Abhijit and Lagun, Dmitry and Laradji, Issam and Liu, Hsueh-Ti (Derek) and Meyer, Henning and Miao, Yishu and Nowrouzezahrai, Derek and Oztireli, Cengiz and Pot, Etienne and Radwan, Noha and Rebain, Daniel and Sabour, Sara and Sajjadi, Mehdi S. M. and Sela, Matan and Sitzmann, Vincent and Stone, Austin and Sun, Deqing and Vora, Suhani and Wang, Ziyu and Wu, Tianhao and Yi, Kwang Moo and Zhong, Fangcheng and Tagliasacchi, Andrea},
    title     = {Kubric: A Scalable Dataset Generator},
    booktitle = CVPR,
    year      = {2022},
}

@inproceedings{zheng2023point,
    author = {Yang Zheng and Adam W. Harley and Bokui Shen and Gordon Wetzstein and Leonidas J. Guibas},
    title = {PointOdyssey: A Large-Scale Synthetic Dataset for Long-Term Point Tracking},
    booktitle = {ICCV},
    year = {2023}
}

@inproceedings{karaev2023dynamicstereo,
  title={DynamicStereo: Consistent Dynamic Depth from Stereo Videos},
  author={Nikita Karaev and Ignacio Rocco and Benjamin Graham and Natalia Neverova and Andrea Vedaldi and Christian Rupprecht},
  booktitle={CVPR},
  year={2023}
}

@inproceedings{Cheng_2022_CVPR,
    author    = {Cheng, Xuelian and Xiong, Huan and Fan, Deng-Ping and Zhong, Yiran and Harandi, Mehrtash and Drummond, Tom and Ge, Zongyuan},
    title     = {Implicit Motion Handling for Video Camouflaged Object Detection},
    booktitle = CVPR,
    year      = {2022},
}

@inproceedings{Liu_2022_CVPR,
    author    = {Liu, Yunze and Liu, Yun and Jiang, Che and Lyu, Kangbo and Wan, Weikang and Shen, Hao and Liang, Boqiang and Fu, Zhoujie and Wang, He and Yi, Li},
    title     = {HOI4D: A 4D Egocentric Dataset for Category-Level Human-Object Interaction},
    booktitle = CVPR,
    year      = {2022},
}

@article{qi2022occluded,
    title={Occluded Video Instance Segmentation: A Benchmark},
    author={Jiyang Qi and Yan Gao and Yao Hu and Xinggang Wang and Xiaoyu Liu and Xiang Bai and Serge Belongie and Alan Yuille and Philip Torr and Song Bai},
    journal=IJCV,
    year={2022},
}

@inproceedings{Li_2019_CVPR,
    author = {Li, Zhengqi and Dekel, Tali and Cole, Forrester and Tucker, Richard and Snavely, Noah and Liu, Ce and Freeman, William T.},
    title = {Learning the Depths of Moving People by Watching Frozen People},
    booktitle = CVPR,
    year = {2019}
}

@inproceedings{FliICCV2013,
author = {Fuxin Li and Taeyoung Kim and Ahmad Humayun and David Tsai and James M. Rehg},
title = { Video Segmentation by Tracking Many Figure-Ground Segments},
booktitle = {ICCV},
year = {2013} 
}

@article{OB14b,
  author       = "P. Ochs and J. Malik and T. Brox",
  title        = "Segmentation of moving objects by long term video analysis",
  journal      = PAMI,
  year         = "2014",
}

@inproceedings{loshchilov2018decoupled,
    title={Decoupled Weight Decay Regularization},
    author={Ilya Loshchilov and Frank Hutter},
    booktitle=ICLR,
    year={2019},
}

@article{oquab2024dinov,
    title={{DINO}v2: Learning Robust Visual Features without Supervision},
    author={Maxime Oquab and Timoth{\'e}e Darcet and Th{\'e}o Moutakanni and Huy V. Vo and Marc Szafraniec and Vasil Khalidov and Pierre Fernandez and Daniel HAZIZA and Francisco Massa and Alaaeldin El-Nouby and Mido Assran and Nicolas Ballas and Wojciech Galuba and Russell Howes and Po-Yao Huang and Shang-Wen Li and Ishan Misra and Michael Rabbat and Vasu Sharma and Gabriel Synnaeve and Hu Xu and Herve Jegou and Julien Mairal and Patrick Labatut and Armand Joulin and Piotr Bojanowski},
    journal={TMLR},
    year={2024},
}

@inproceedings{ryali2023hiera,
  title={Hiera: A Hierarchical Vision Transformer without the Bells-and-Whistles},
  author={Ryali, Chaitanya and Hu, Yuan-Ting and Bolya, Daniel and Wei, Chen and Fan, Haoqi and Huang, Po-Yao and Aggarwal, Vaibhav and Chowdhury, Arkabandhu and Poursaeed, Omid and Hoffman, Judy and Malik, Jitendra and Li, Yanghao and Feichtenhofer, Christoph},
  booktitle={ICML},
  year={2023}
}

@inproceedings{cheng2021maskformer,
  title={Per-Pixel Classification is Not All You Need for Semantic Segmentation},
  author={Bowen Cheng and Alexander G. Schwing and Alexander Kirillov},
  booktitle=NIPS,
  year={2021}
}

@inproceedings{Li_2023_CVPR,
    author    = {Li, Feng and Zhang, Hao and Xu, Huaizhe and Liu, Shilong and Zhang, Lei and Ni, Lionel M. and Shum, Heung-Yeung},
    title     = {Mask DINO: Towards a Unified Transformer-Based Framework for Object Detection and Segmentation},
    booktitle = CVPR,
    year      = {2023},
}

@article{Nasaruddin2020,
  author   = {Nasaruddin, Nasaruddin and Muchtar, Kahlil and Afdhal, Afdhal and Dwiyantoro, Alvin Prayuda Juniarta},
  title    = {Deep anomaly detection through visual attention in surveillance videos},
  journal  = {Journal of Big Data},
  year     = {2020},
}

@inproceedings{chen2021lidarmos,
  author    = {Chen, Xieyuanli and Li, Shijie and Mersch, Benedikt and Wiesmann, Louis and Gall, J{\"u}rgen and Behley, Jens and Stachniss, Cyrill},
  title     = {Moving Object Segmentation in {3D} {LiDAR} Data: A Learning-based Approach Exploiting Sequential Data},
  booktitle = {IROS},
  year      = {2021},
}

@article{li2023mosfusion,
  author  = {Li, Qipeng and Zhuang, Yuan and Chen, You and Huai, Jianzhu and Li, Miaomiao and Ma, Tianxiang and Tang, Yufei and Liang, Xinlian},
  title   = {Multi-sensor Fusion for Robust Localization with Moving Object Segmentation in Complex Dynamic {3D} Scenes},
  journal = {International Journal of Applied Earth Observation and Geoinformation},
  year    = {2023},
}

@article{weng2021aten,
  author  = {Weng, Zhengkui and Jin, Zhipeng and Chen, Shuangxi and Shen, Quanquan and Ren, Xiangyang and Li, Wuzhao},
  title   = {Attention-Based Temporal Encoding Network with Background-Independent Motion Mask for Action Recognition},
  journal = {Computational Intelligence and Neuroscience},
  year    = {2021},
}

@inproceedings{lin2026depth,
    title={Depth Anything 3: Recovering the Visual Space from Any Views},
    author={Haotong Lin and Sili Chen and Jun Hao Liew and Donny Y. Chen and Zhenyu Li and Yang Zhao and Sida Peng and Hengkai Guo and Xiaowei Zhou and Guang Shi and Jiashi Feng and Bingyi Kang},
    booktitle=ICLR,
    year={2026},
}

@InProceedings{Choudhury22,
  author       = "Subhabrata Choudhury and Laurynas Karazija and Iro Laina and Andrea Vedaldi and Christian Rupprecht",
  title        = "{G}uess {W}hat {M}oves: {U}nsupervised {V}ideo and {I}mage {S}egmentation by {A}nticipating {M}otion",
  booktitle    = bmvc,
  year         = "2022",
}

@InProceedings{Dutta19a,
  author       = "Abhishek Dutta and Andrew Zisserman",
  title        = "The VIA Annotation Software for Images, Audio and Video",
  booktitle    = acmmm,
  year         = "2019",
}

@InProceedings{Karaev25,
  author       = "Nikita Karaev and Yuri Makarov and Jianyuan Wang and Natalia Neverova and Andrea Vedaldi and Christian Rupprecht",
  title        = "CoTracker3: Simpler and Better Point Tracking by Pseudo-Labeling Real Videos",
  booktitle    = iccv,
  year         = "2025",
}

@InProceedings{Karazija24b,
  author       = "Laurynas Karazija and Iro Laina and Christian Rupprecht and Andrea Vedaldi",
  title        = "Learning segmentation from point trajectories",
  booktitle    = neurips,
  year         = "2024",
}

@InProceedings{Lamdouar20,
  author       = "Hala Lamdouar and Charig Yang and Weidi Xie and Andrew Zisserman",
  title        = "Betrayed by Motion: Camouflaged Object Discovery via Motion Segmentation",
  booktitle    = accv,
  year         = "2020",
}

@InProceedings{Wang19,
  author       = "Jiangliu Wang and Jianbo Jiao and Linchao Bao and Shengfeng He and Yunhui Liu and Wei Liu",
  title        = "Self-Supervised Spatio-Temporal Representation Learning for Videos by Predicting Motion and Appearance Statistics",
  booktitle    = cvpr,
  year         = "2019",
}

@InProceedings{Wang25,
  author       = "Jianyuan Wang and Minghao Chen and Nikita Karaev and Andrea Vedaldi and Christian Rupprecht and David Novotny",
  title        = "VGGT: Visual Geometry Grounded Transformer",
  booktitle    = cvpr,
  year         = "2025",
}

@InProceedings{Xie22,
  author       = "Junyu Xie and Weidi Xie and Andrew Zisserman",
  title        = "Segmenting Moving Objects via an Object-Centric Layered Representation",
  booktitle    = nips,
  year         = "2022",
}

@InProceedings{Xie24,
  author       = "Junyu Xie and Weidi Xie and Andrew Zisserman",
  title        = "Appearance-Based Refinement for Object-Centric Motion Segmentation",
  booktitle    = eccv,
  year         = "2024",
}

@InProceedings{Xie24a,
  author       = "Junyu Xie and Charig Yang and Weidi Xie and Andrew Zisserman",
  title        = "Moving Object Segmentation: All You Need Is SAM (and Flow)",
  booktitle    = accv,
  year         = "2024",
}

@InProceedings{Yang21a,
  author       = "Charig Yang and Hala Lamdouar and Erika Lu and Andrew Zisserman and Weidi Xie",
  title        = "Self-supervised Video Object Segmentation by Motion Grouping",
  booktitle    = iccv,
  year         = "2021",
}

\newpage
\appendix
\renewcommand{\thetable}{\thesection\arabic{table}}
\renewcommand{\thefigure}{\thesection\arabic{figure}}
\renewcommand{\theequation}{\thesection\arabic{equation}}
\setcounter{table}{0}
\setcounter{figure}{0}
\setcounter{equation}{0}
\section*{Appendix}
\noindent This appendix is organised as follows:
\vspace{0.1cm}
\begin{itemize}[leftmargin=1.2em, itemindent=1pt, itemsep=2pt, parsep=2pt, topsep=1pt]
    \item \textbf{Additional method details:} Appendix~\ref{appsec:method} describes the architecture and training objective of the \method-S variant, and provides the full pseudo-code and elementary-step walkthrough for the \method propagator.
    \item \textbf{Additional implementation details:} Appendix~\ref{appsec:imple} reports the \method-S training setup, the propagator thresholds, and the protocol for runtime measurement.
    \item \textbf{Dataset details:} Appendix~\ref{appsec:dataset} provides per-dataset statistics for our training and evaluation sources, along with example visualisations from \method-2K.
    \item \textbf{\method-2K curation details:} Appendix~\ref{appsec:gmos_curation} details the annotator instructions, the annotation tool, and per-subset curation statistics together with the annotation cost.
    \item \textbf{Ablation study:} Appendix~\ref{appsec:abla} presents comprehensive ablations on both the \method proposer and the \method propagator, alongside repeated training runs across random seeds to assess statistical stability.
    \item \textbf{Additional qualitative visualisations:} Appendix~\ref{appsec:qual} provides further qualitative comparisons with baselines and additional in-the-wild MOS/MOS-I results.
    \item \textbf{Discussion:} Appendix~\ref{appsec:disc} discusses current failure cases, outlines possible extensions, and addresses broader societal impact.
\end{itemize}
We additionally release, in the supplementary material, (i) a \emph{visualisation video} covering the evaluation protocol comparison (MOS vs.\ MOS-I), dataset samples, and model outputs, and (ii) a \emph{code repository} with training and evaluation pipelines, along with temporally fine-grained motion state labels for the \method-2K subsets, indexed by dataset, split, and sequence name.
\vspace{8pt}

\section{Additional method details}
\label{appsec:method}

\subsection{\method-S}
\label{appsubsec:gmos-s}
\begin{figure*}[h]
\centering
\vspace{3mm}
\includegraphics[width=0.7\linewidth]{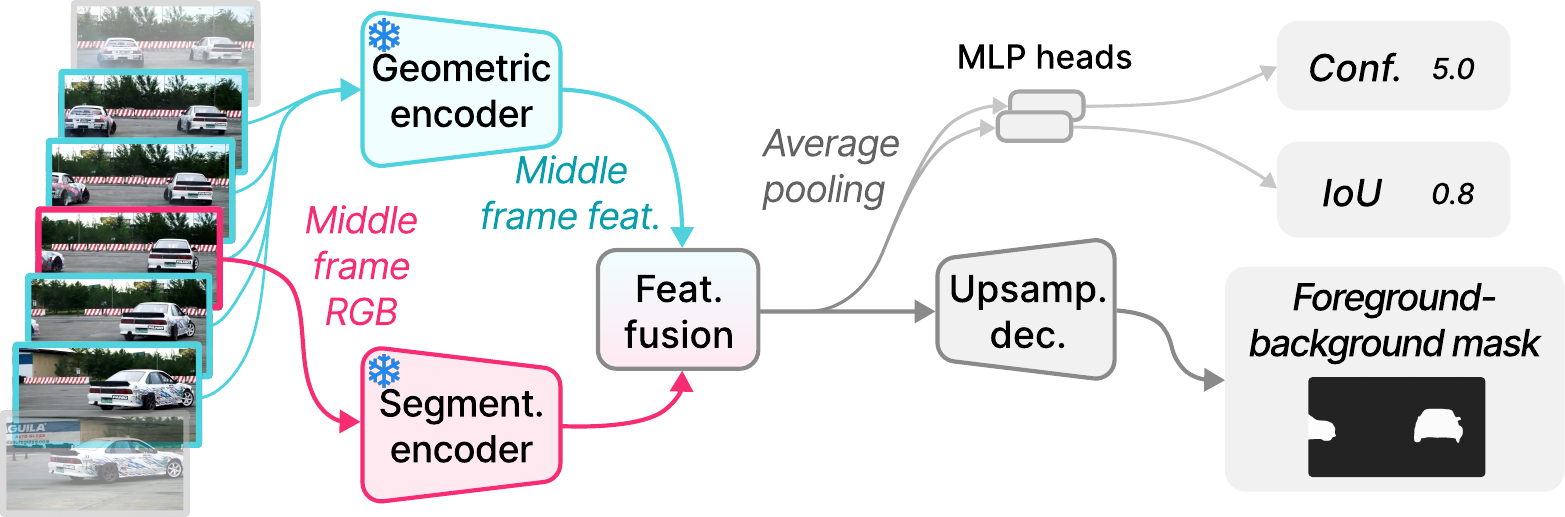}
\vspace{1mm}
\caption{
\textbf{Overview of \method-S,} a streaming foreground--background variant of the \method proposer. Instead of query-based decoding, \method-S directly upsamples dense features to predict a binary mask.
}
\vspace{1mm}
\label{fig:model_s}
\end{figure*}

\noindentpara{Architecture.}
\method-S is designed as a streaming foreground--background variant of the \method proposer. As illustrated in \Cref{fig:model_s}, it shares the frozen geometric encoder ($\pi^3$), the frozen segmentation encoder (SAM2), and the feature fusion module with the \method proposer (\Cref{fig:model}). Two design choices distinguish it from the full proposer: (\textit{i})~the transformer decoder and the $N$ learnable object queries are removed, and the fused feature is passed directly through an upsampling decoder that predicts a single binary foreground--background mask per frame; and (\textit{ii})~since motion is implicit in the union of moving objects, the motion head is dropped, and the remaining confidence and IoU heads operate on globally average-pooled fused features, producing one scalar per frame instead of one per object. 

In addition, we dispense with the propagator, as no object-level proposals are produced by \method-S. Predictions are therefore instantaneous and online, making \method-S the fastest variant in our framework.

\noindentpara{Training objective.}
\method-S predicts a single frame-level foreground--background mask $\hat{M}_t$, so the per-object Hungarian matching and motion loss used in the \method proposer are no longer needed. We therefore simplify \cref{eq:loss} into
\begin{equation}
\mathcal{L}^{\text{S}} = \lambda_{\text{iou}}^{\text{S}}\mathcal{L}^{\text{S}}_{\text{iou}} + \tilde{c}\, \lambda_{\text{mask}}^{\text{S}}\mathcal{L}^{\text{S}}_{\text{mask}} - \lambda_{\text{conf}}^{\text{S}}\log\tilde{c},
\label{eq:loss_s}
\end{equation}
where {\small $\mathcal{L}^{\text{S}}_{\text{mask}}$} and {\small $\mathcal{L}^{\text{S}}_{\text{iou}}$} are the frame-level counterparts of the proposer's mask and IoU losses in~\cref{eq:loss}, and {\small $\tilde{c} \in [1, c_{\text{max}}]$} is the rescaled frame-level confidence. Since \method-S reasons at the frame level rather than the object level, the split confidence regulariser of the \method proposer collapses to a single term.

\subsection{\method propagator}
\label{appsubsec:gmos-prop}

\begin{algorithm}[!pt]
\caption{Pseudo-code of the \method propagator.}
\label{alg:gmos}
\begin{algorithmic}[1]
\LineCommentBase{Thresholds (values in Appendix~\ref{appsec:imple}):}
\LineCommentBase{\;\;\;$\tau_m$: an object is moving if its motion score exceeds this.}
\LineCommentBase{\;\;\;$\tau_u$: the proposer IoU prediction is trusted if it exceeds this.}
\LineCommentBase{\;\;\;$\tau_{\text{match}}$: proposer and propagated masks refer to the same track if their IoU exceeds this.}
\LineCommentBase{\;\;\;$\tau_{\text{new}}$: a proposer mask is treated as a new track if its max precision w.r.t. any propagated mask is below this.}
\Statex
\LineCommentBase{Helper functions:}
\LineCommentBase{\;\;\;$\Call{Precision}{A, B} = |A \cap B|/|A|$: the fraction of mask $A$ covered by mask $B$.}
\LineCommentBase{\;\;\;\textsc{HungarianMatch}$(\tilde{M}_t, \{\hat{M}_t, \hat{m}_t, \hat{u}_t\})$: solve a bipartite assignment between propagated masks $\tilde{M}_t$ and proposer masks $\hat{M}_t$ on mask IoU, returning the proposer predictions reordered to align with the propagated tracks.}
\Statex
\LineCommentBase{SAM2 operations:}
\LineCommentBase{\;\;\;\textsc{InitSAM2}$(\{(t,i,M^{(i)}_t)\})$: initialise a SAM2 state $\mathcal{S}$ with a set of mask prompts, each tagging object $i$ at frame $t$ with mask $M^{(i)}_t$.}
\LineCommentBase{\;\;\;\textsc{SAM2Propagate}$(\mathcal{S}, t)$: propagate the tracks currently held in $\mathcal{S}$ to frame $t$, returning the propagated masks $\tilde{M}_t$.}
\LineCommentBase{\;\;\;\textsc{AddPrompt}$(\mathcal{S}, t, i, M)$: inject mask $M$ as an additional prompt for object $i$ at frame $t$ into $\mathcal{S}$.}
\Statex
\LineCommentBase{Elementary step}
\algrenewcommand\algorithmicprocedure{\textbf{step}}
\Procedure{PropStep}{$\mathcal{S}, t_{\text{start}}, \text{direction}, \text{update\_prompt}$}
    \ColorComment{SAM2 state $\mathcal{S}$, starting frame, propagation direction $\in\{+1,-1\}$, whether to update the prompt set during propagation}
    \For{$t \gets t_{\text{start}}$, stepping by direction}
        \State $\tilde{M}_t \gets \Call{SAM2Propagate}{\mathcal{S}, t}$
        \ColorComment{propagated masks at frame $t$}
        \State $\{\hat{M}_t, \hat{m}_t, \hat{u}_t\} \gets \Call{Proposer}{I_{t-n},\dots,I_{t+n}}$
        \ColorComment{mask, motion, IoU predictions}
        \State $\{\hat{M}_t, \hat{m}_t, \hat{u}_t\} \gets \Call{HungarianMatch}{\tilde{M}_t, \{\hat{M}_t, \hat{m}_t, \hat{u}_t\}}$
        \If{update\_prompt}
            \For{each object $i$ with $\hat{m}^{(i)}_t > \tau_m$ \textbf{and} $\hat{u}^{(i)}_t > \tau_u$}
                \If{$\max_{j}\Call{Precision}{\hat{M}^{(i)}_t, \tilde{M}^{(j)}_t} < \tau_{\text{new}}$}
                    \State $\mathcal{S} \gets \Call{AddPrompt}{\mathcal{S}, t, i_{\text{new}}, \hat{M}^{(i)}_t}$
                    \ColorComment{\textit{add} new track}
                \ElsIf{$\Call{IoU}{\hat{M}^{(i)}_t, \tilde{M}^{(i)}_t} > \tau_{\text{match}}$}
                    \State $\mathcal{S} \gets \Call{AddPrompt}{\mathcal{S}, t, i, \hat{M}^{(i)}_t}$
                    \ColorComment{\textit{reinforce} existing track $i$}
                \EndIf
            \EndFor
        \EndIf
        \State $\tilde{m}^{(i)}_t \gets \mathbf{1}\!\left[\hat{m}^{(i)}_t > \tau_m\right]$ \ColorComment{motion label; tracks without a moving match are labelled static}
        \State \textbf{store} $(\tilde{M}^{(i)}_t, \tilde{m}^{(i)}_t, \tilde{u}^{(i)}_t)$ for all $i$
    \EndFor
    \State \Return $\mathcal{S}$, stored tracks
\EndProcedure
\algrenewcommand\algorithmicprocedure{\textbf{procedure}}
\Statex
\LineCommentBase{Online procedure: a single forward pass with injection}
\Procedure{Online}{}
    \State $\mathcal{S} \gets \Call{InitSAM2}{\{(1,i,\hat{M}^{(i)}_1) : \hat{m}^{(i)}_1 > \tau_m,\, \hat{u}^{(i)}_1 > \tau_u\}}$
    \State $\mathcal{S}, \mathcal{T}^{\text{on}} \gets \Call{PropStep}{\mathcal{S}, 1, +1, \text{update\_prompt}=\text{True}}$
    \State \Return $\mathcal{T}^{\text{on}}$
\EndProcedure
\Statex
\LineCommentBase{Offline procedure: refines online output with a bi-directional second pass}
\Procedure{Offline}{}
    \State $\mathcal{T}^{\text{on}} \gets \Call{Online}{}$
    \State $P \gets \Call{SelectTopK}{\mathcal{T}^{\text{on}}}$ \ColorComment{per-object top-K high-quality masks}
    \State $\mathcal{S} \gets \Call{InitSAM2}{P}$
    \State $t^\star \gets \arg\max_t \sum_{i} \tilde{u}^{\text{on},(i)}_{t}\,\mathbf{1}[\tilde{u}^{\text{on},(i)}_{t} > \tau_u]$ \ColorComment{anchor frame from online IoU scores}
    \State $\mathcal{S}, \mathcal{T}^{\text{off}}_{+} \gets \Call{PropStep}{\mathcal{S}, t^\star, +1, \text{update\_prompt}=\text{False}}$
    \State $\mathcal{S}, \mathcal{T}^{\text{off}}_{-} \gets \Call{PropStep}{\mathcal{S}, t^\star, -1, \text{update\_prompt}=\text{False}}$
    \State \Return $\mathcal{T}^{\text{off}}_{+} \cup \mathcal{T}^{\text{off}}_{-}$
\EndProcedure
\end{algorithmic}
\end{algorithm}

We now expand on the elementary propagation step and the online/offline procedures outlined in Sec.~\ref{subsec:propagator} and provide the full pseudo-code in \Cref{alg:gmos}. The propagator maintains a set of prompts in the SAM2 state $\mathcal{S}$ and produces per-frame masks {\small $\tilde{M}_t$} and motion labels {\small $\tilde{m}_t$}.

\noindentpara{\textsc{PropStep}.} Starting from {\small $t_{\text{start}}$} and moving in the given direction, for each frame $t$:
\begin{enumerate}[leftmargin=1.6em, itemindent=1pt, itemsep=2pt, parsep=2pt, topsep=1pt, label=(\textit{\roman*})]
    \item SAM2 propagates the currently tracked masks to frame $t$, yielding {\small $\tilde{M}_t$}.
    \item The proposer runs on a short local window to obtain predictions {\small $\{\hat{M}_t, \hat{m}_t, \hat{u}_t\}$} (\ie mask proposals, motion probabilities, and estimated IoUs), which are Hungarian-matched to {\small $\tilde{M}_t$} on mask IoU.
    \item If $\text{update\_prompt}$ is enabled, the prompt set is updated by one of two rules applied only to confident moving proposals ({\small $\hat{m}^{(i)}_t > \tau_m$}, {\small $\hat{u}^{(i)}_t > \tau_u$}):
    (i) \emph{addition}, when the proposer mask is not covered by any existing track ({\small $\max_j \Call{Precision}{\hat{M}^{(i)}_t, \tilde{M}^{(j)}_t} < \tau_{\text{new}}$}), seeds a new track;
    (ii) \emph{reinforcement}, when the matched proposer mask aligns with the propagated mask ({\small $\Call{IoU}{\hat{M}^{(i)}_t, \tilde{M}^{(i)}_t} > \tau_{\text{match}}$}), injects {\small $\hat{M}^{(i)}_t$} as an additional prompt for the same track.
    \item The per-frame motion label {\small $\tilde{m}^{(i)}_t$} is inherited from the matched proposer prediction, {\small $\tilde{m}^{(i)}_t = \mathbf{1}[\hat{m}^{(i)}_t > \tau_m]$}. Existing tracks for which no confident moving proposer prediction matches at frame $t$ (either unmatched or matched with {\small $\hat{m}^{(i)}_t \leq \tau_m$}) are therefore labelled static, capturing the ``moving-then-static'' transition.
\end{enumerate}

\noindentpara{Online procedure.} We initialise $\mathcal{S}$ with confident moving proposals from frame~$1$ (\ie the first frame) and run \textsc{PropStep} once in the forward direction with $\text{update\_prompt}=\text{True}$. The procedure is causal and streams per-frame outputs as the video arrives, making it directly suitable for MOS-I deployment.

\noindentpara{Offline procedure.} We first run the online procedure, then select the top-$K$ highest-quality masks per object from the resulting tracks as a fixed prompt set, and re-initialise $\mathcal{S}$ with it. We pick an anchor frame {\small $t^\star = \arg\max_t \sum_i \tilde{u}^{\text{on},(i)}_t\,\mathbf{1}[\tilde{u}^{\text{on},(i)}_t > \tau_u]$} from the online per-object IoU scores, and run \textsc{PropStep} forward and backward from {\small $t^\star$} with $\text{update\_prompt}=\text{False}$. The curated prompt set and bi-directional propagation together recover masks on frames where the online pass missed an object or produced a low-quality mask.

\section{Additional implementation details}
\label{appsec:imple}

\para{\method-S training setup.}
\method-S follows the same optimiser, schedule, and hardware as the \method proposer (Sec.~\ref{subsec:impl}). For the loss in~\cref{eq:loss_s}, we set {\small $\lambda^\text{S}_\text{mask}{=}20$}, {\small $\lambda^\text{S}_\text{iou}{=}1$}, and {\small $\lambda_\text{conf}^\text{S}{=}0.5$}, with a rescaled confidence upper bound {\small $c_{\text{max}} = 5$}. The frame-level mask loss combines focal loss ($\alpha{=}0.25$, $\gamma{=}2$) and dice loss in a $20{:}1$ ratio.

\para{Propagator settings.}
For the thresholds used in \Cref{alg:gmos}, we set $\tau_m = 0.5$ (motion score above which an object is considered moving), $\tau_u = 0.7$ (proposer IoU prediction above which a mask is trusted), $\tau_{\text{match}} = 0.95$ (IoU between the proposer and propagated masks above which the proposer mask reinforces the existing track), and $\tau_{\text{new}} = 0.3$ (maximum per-track precision below which the proposer mask is treated as a new object). For the offline \textsc{SelectTopK} step, we pick the top $10\%$ of frames per object ranked by proposer IoU prediction, keeping only those with IoU above $0.95$.

\para{Runtime measurement.}
Per-frame runtimes in \Cref{tab:method-comparison} are measured on a single NVIDIA RTX A6000 GPU over all frames of DAVIS17. Each measurement includes preparation of auxiliary input modalities (optical flow, depth, or point tracks), the model forward pass, and any post-processing (\eg SAM2 refinement), but excludes data-loading overhead.

\section{Dataset details}
\label{appsec:dataset}

\para{Overall dataset statistics.}
As described in \Cref{sec:dataset}, our training set comprises three synthetic datasets (Kubric, PointOdyssey, and DynamicReplica), five real-video subsets from \method-2K, and the Mannequin Challenge dataset~\cite{Li_2019_CVPR}, which contains no moving objects and serves as a source of static-scene negatives. For temporally fine-grained MOS evaluation, we use the DAVIS17 and YTVOS19 subsets of \method-2K as test sets. All datasets are open-sourced to research usage, with licence and terms in the datasets strictly respected.

\Cref{tab:dataset-stats-full} reports the full statistics of each dataset, including the number of videos, total duration, annotated frames, object count, and the proportion of frames containing motion. Subsets belonging to \method-2K are marked with an asterisk~(*). In total, the training set spans $14{,}171$ videos with $782{,}128$ annotated frames and $34{,}506$ objects.
\begin{table*}[t]
\centering
\caption{\textbf{Overall dataset statistics.} Asterisks (*) denote subsets in \method-2K.}
\vspace{-1mm}
\label{tab:dataset-stats-full}
\setlength{\tabcolsep}{4pt}
\resizebox{\linewidth}{!}{%
\begin{tabular}{llrrrrrrr}
\toprule
Split & Dataset & Videos & \makecell[r]{Duration\\ (hr)} & Anno. frames & \makecell[r]{Anno. frames\\/ video} & Objects & \makecell[r]{Objects \\/ video} & \makecell[r]{Motion prop. \\ /object} \\
\midrule
\multirow{3}{*}{\makecell[l]{Train\\(Synthetic)}}
  & Kubric~\cite{Greff_2022_CVPR}              & 10{,}000 & 5.56  & 240{,}000 & 24.0    & 29{,}294 & 2.93 & 60.8\% \\
  & PointOdyssey~\cite{zheng2023point}        & 43       & 1.24  & 107{,}514 & 2500.0 & 64       & 1.49 & 57.1\% \\
  & DynamicReplica~\cite{karaev2023dynamicstereo}      & 483      & 1.68  & 144{,}900 & 300.0   & 933      & 1.93 & 60.7\% \\
\midrule
\multirow{6}{*}{\makecell[l]{Train\\(Real)}}
  & *DAVIS17~\cite{Ponttuset17}   & 44       & 0.03  & 3{,}101   & 70.5  & 92       & 2.09 & 94.8\% \\
  & *MoCA-Mask~\cite{Cheng_2022_CVPR}          & 26       & 0.11  & 1{,}978   & 76.1  & 26       & 1.00 & 54.8\% \\
  & *HOI4D~\cite{Liu_2022_CVPR}            & 273      & 0.95  & 81{,}900  & 300.0 & 609      & 2.23 & 69.9\% \\
  & *OVIS~\cite{qi2022occluded}              & 404      & 0.24  & 25{,}925  & 64.2  & 1{,}865  & 4.62 & 92.6\% \\
  & *YTVOS19~\cite{Xu18}              & 1{,}183  & 1.74  & 31{,}327  & 26.5  & 1{,}623  & 1.37 & 91.1\% \\
  & Mannequin Challenge~\cite{Li_2019_CVPR} & 1{,}715  & 1.22  & 145{,}483 & 84.8  & 0        & 0.00 & 0.0\%  \\
\midrule
\multicolumn{2}{l}{\textbf{Total (Train)}} & \textbf{14{,}171} & \textbf{12.77} & \textbf{782{,}128} & \textbf{55.2} & \textbf{34{,}506} & \textbf{2.43} & \textbf{64.2\%} \\
\midrule
\multirow{2}{*}{Test}
  & *DAVIS17~\cite{Ponttuset17}   & 19  & 0.01 & 1{,}237 & 65.1 & 31  & 1.63 & 96.9\% \\
  & *YTVOS19~\cite{Xu18}  & 261 & 0.40 & 7{,}115 & 27.3 & 402 & 1.54 & 91.5\% \\
\midrule
\multicolumn{2}{l}{\textbf{Total (Test)}} & \textbf{280} & \textbf{0.41} & \textbf{8{,}352} & \textbf{29.8} & \textbf{433} & \textbf{1.55} & \textbf{91.9\%} \\
\bottomrule
\end{tabular}%
}
\end{table*}

\begin{figure*}[t]
\includegraphics[width=\linewidth, trim=0 0 0 0, clip]{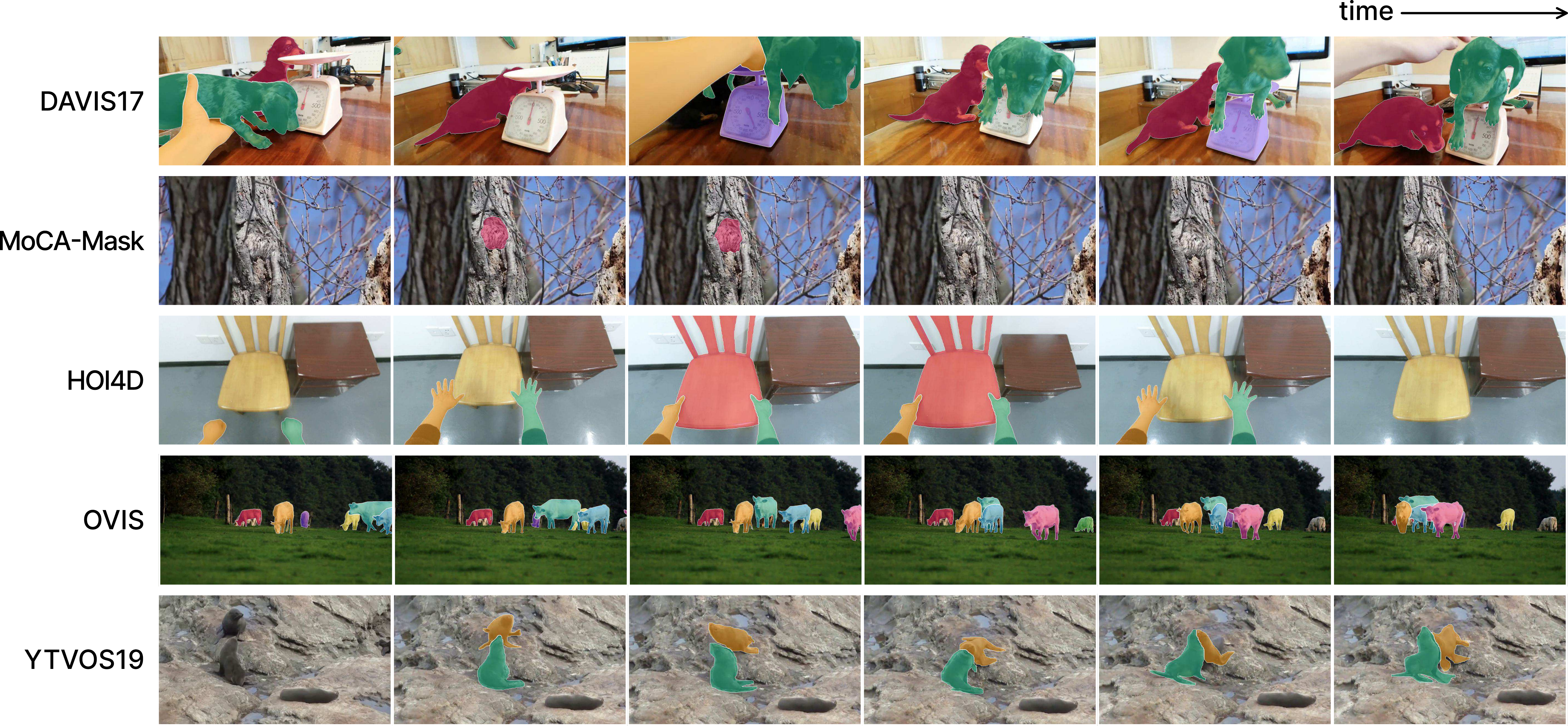}
\vspace{-4mm}
\caption{
\textbf{\method-2K example sequences.} 
Videos are sampled from each of the five constituent subsets, with ground-truth segmentation masks overlaid. Frames are sampled uniformly along the temporal axis (left to right). An object is masked only in frames where it is actively moving, reflecting the temporally fine-grained motion labels in \method-2K.
}
\label{fig:gmos2k_vis}
\end{figure*}

\vspace{1mm}
\para{Example visualisations in \method-2K.} \Cref{fig:gmos2k_vis} presents example sequences from each of the five \method-2K subsets, illustrating the diversity of scenes and motion patterns. An object is masked only in frames where it is actively moving, reflecting the temporally fine-grained nature of annotated motion labels.
For instance, in HOI4D, two hands are in motion throughout most of the sequence, whereas the chair is masked only during the brief interval when it is being picked up. Similarly, the YTVOS19 example shows that the seal in the bottom-right corner is never masked, as it remains stationary throughout the sequence.

\section{\method-2K curation details}
\label{appsec:gmos_curation}

\subsection{Instructions and tools for annotators}

\noindentpara{Video-level filtering.}
As described in \Cref{sec:dataset}, the curation pipeline begins by filtering $5{,}001$ candidate videos through two sequential questions: (\textit{i})~whether all independently moving objects have ground-truth segmentation masks, and (\textit{ii})~whether all annotated objects are moving throughout the entire sequence. Annotators review each video and label their answers accordingly. The detailed instructions provided to annotators are shown below.

\begin{tcolorbox}[
    enhanced,
    breakable,
    colback=gray!5,
    colframe=gray!20,
    title=\!\!\!\!\!\textbf{Video-level filtering instructions},
    fonttitle=\bfseries\fontsize{9.5}{11}\selectfont,
    coltitle=black,
    arc=2mm,
    boxrule=0.6pt
]
\small

For each video, please answer the following two questions:

\vspace{0.4em}
\textbf{Question 1: Do all moving objects in the video have corresponding segmentation masks?}
\begin{itemize}\itemsep0.2em
    \item You may ignore all stationary objects (only moving objects need to be checked).
    \item \textbf{Yes} $\rightarrow$ label as \texttt{1}
    \item \textbf{No} $\rightarrow$ label as \texttt{-1}
\end{itemize}

\textbf{Question 2: Are all segmented objects moving in every single frame of the video?}
\begin{itemize}\itemsep0.2em
    \item \textbf{Yes} (every segmented object is moving in every frame) $\rightarrow$ label as \texttt{1}
    \item \textbf{No} (at least one segmented object is stationary in some frame, or all are stationary) $\rightarrow$ label as \texttt{-1}
\end{itemize}

\textit{Note: If Question 1 is labeled \texttt{-1}, you may skip Question 2.}

\end{tcolorbox}

\begin{figure*}[t]
\includegraphics[width=1.0\linewidth]{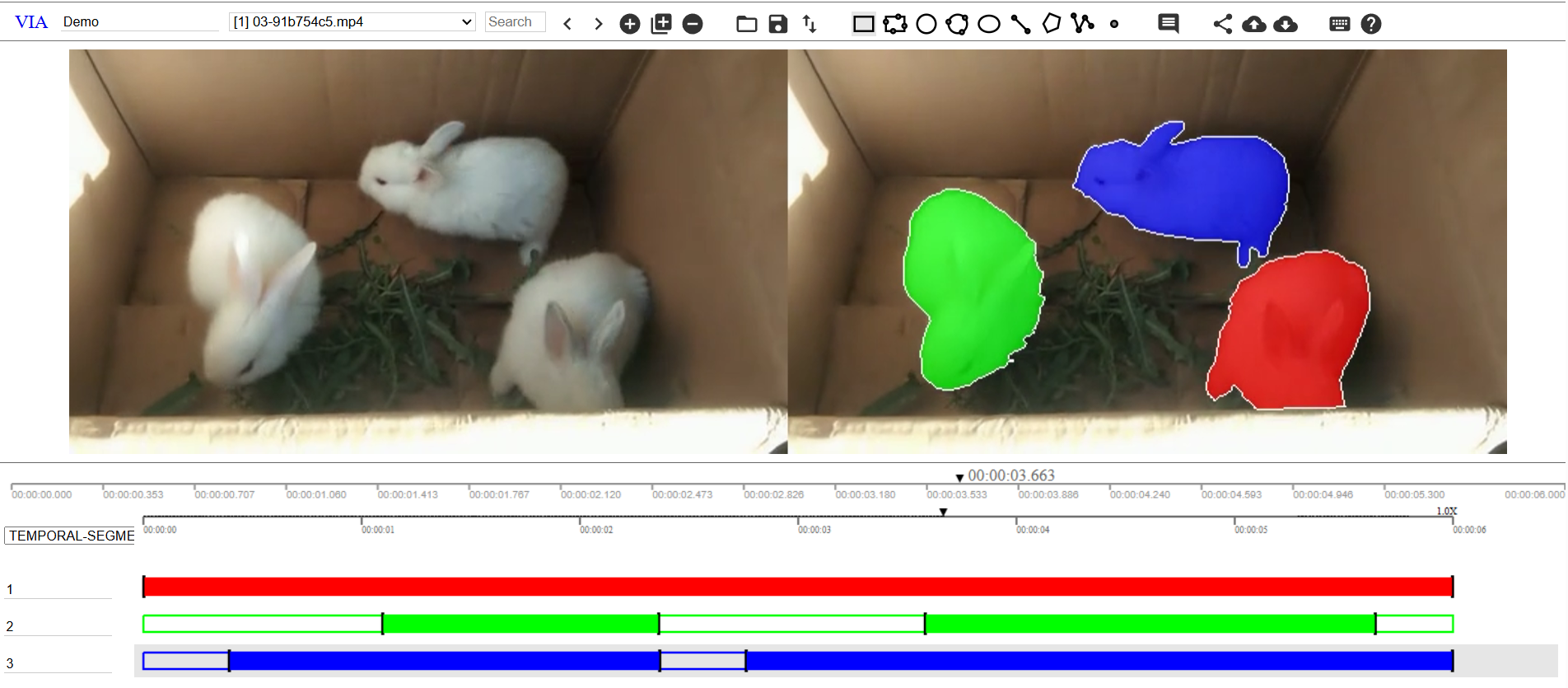}
\caption{
\textbf{VIA interface for Temporally Fine-grained Annotation (TFA).}}
\label{fig:via}
\end{figure*}

\vspace{2mm}
\para{Temporally Fine-grained Annotation (TFA).}
Videos that pass the first filter but fail the second are forwarded for TFA, which labels per-object motion intervals along the temporal axis. We perform TFA using VIA~\cite{Dutta19a}, an open-source annotation tool with built-in temporal grounding functionality. As shown in \Cref{fig:via}, we adapt the default VIA interface with two modifications: (\textit{i})~the original video and its mask-overlaid counterpart are displayed side-by-side, with each object assigned a unique colour for easy identification; and (\textit{ii})~the temporal axes are colour-coded to correspond to individual objects. The detailed annotation workflow and instructions are provided below.

\begin{tcolorbox}[
    enhanced,
    breakable,
    colback=gray!5,
    colframe=gray!20,
    title=\textbf{\!\!\!\!\!Temporally Fine-grained Annotation (TFA) instructions},
    fonttitle=\bfseries\fontsize{9.5}{11}\selectfont,
    coltitle=black,
    arc=2mm,
    boxrule=0.6pt
]
\small
(\Cref{fig:via} provides an example VIA interface)

\vspace{0.3em}
\textbf{Step 1.} Enter the VIA annotation tool and import a video to annotate.

\vspace{0.3em}
\textbf{Step 2.} \textit{(Optional)} Open ``Keyboard Shortcuts'' to learn VIA operations.

\vspace{0.3em}
\textbf{Step 3.} Each object is associated with an automatically generated time axis identified by colour codes. For each object, label the temporal windows when the object \textit{truly moves} (regardless of whether the camera is moving).

\begin{itemize}\itemsep0.2em
    \item Only focus on objects with segmentation masks; ignore motions of shadows, smoke, fluid, etc.
    \item When the object is invisible, treat it as ``static'' --- do not label.
    \item If the object moves all the time, label the full time axis. If the object is always static or invisible, leave its time axis empty.
\end{itemize}

\textbf{Step 4.} Click the ``Export'' button and save the CSV file.

\vspace{0.3em}
\textbf{Step 5.} Return to Step 1.

\vspace{0.4em}
\textit{Note: Step 4 (saving) can be done for a single video or multiple videos together.}

\end{tcolorbox}

\vspace{2mm}
\para{Quality control.}
We apply quality control protocols at each of the above stages.
\begin{itemize}[leftmargin=1.2em, itemsep=2pt, parsep=1pt, topsep=0pt]
    \item \textit{Video-level filtering.} Two annotators independently answer the two filtering questions for each candidate video. Sequences on which they disagree are flagged and jointly reviewed until a consensus is reached, ensuring that no qualifying video is dropped and no unsuitable video is forwarded to TFA.
    \item \textit{Temporally Fine-grained Annotation (TFA).} After each sequence is annotated, a second annotator visualises the resulting MOS-I ground truth, with each object's mask overlaid only on the frames within its labelled motion intervals. The reviewer checks for (i) moving objects whose motion intervals are missing entirely (an object never receives a mask despite moving), (ii) static or background objects falsely labelled as moving (a mask appears on frames where the object is at rest), and (iii) inaccurate motion--rest transition boundaries. This step also captures errors from the filtering stage, removing any unsuitable videos that erroneously passed the previous filter. Flagged cases are returned to the original annotator for revision and re-verified by the reviewer.
\end{itemize}

\subsection{Curation statistics}

\noindentpara{Per-subset statistics.}
\Cref{tab:gmos2k_curation} provides a detailed breakdown of the curation pipeline described in \Cref{sec:dataset}, reporting the number of videos retained after each filtering and annotation stage for every \method-2K subset. Of the $5{,}001$ initial candidate videos, $2{,}791$ are discarded by the first video-level filter, and the remaining $2{,}210$ form the final dataset, of which $1{,}467$ require TFA.

\begin{table}[th]
\centering
\caption{\textbf{Curation statistics for \method-2K.} ``TFA'' stands for Temporally Fine-grained Annotations.}
\vspace{1.5mm}
\label{tab:gmos2k_curation}
\setlength{\tabcolsep}{8pt}
\renewcommand{\arraystretch}{1.15}
\resizebox{0.8\linewidth}{!}{%
\begin{tabular}{llrrrrr}
\toprule
\multirow{2}{*}{Split} & \multirow{2}{*}{Dataset} & \multirow{2}{*}{\makecell[r]{Initial\\videos}} & \multirow{2}{*}{\makecell[r]{Filtered\\videos}} & \multicolumn{3}{c}{Videos in \method-2K} \\
\cmidrule(lr){5-7}
 & & & & w/o TFA & w/ TFA & Total \\
\midrule
\multirow{5}{*}{Train}
 & DAVIS17~\cite{Ponttuset17}       & 60      & 16      & 32   & 12   & 44      \\
 & MoCA-Mask~\cite{Cheng_2022_CVPR}   & 26      & 0       & 2    & 24   & 26      \\
 & HOI4D~\cite{Liu_2022_CVPR}       & 300     & 27      & 0    & 273  & 273     \\
 & OVIS~\cite{qi2022occluded}        & 607     & 203     & 59   & 345  & 404     \\
 & YTVOS19~\cite{Xu18}         & 3{,}471 & 2{,}288 & 526  & 657  & 1{,}183 \\
\midrule
\textbf{Total (Train)} & & \textbf{4{,}464} & \textbf{2{,}534} & \textbf{619} & \textbf{1{,}311} & \textbf{1{,}930} \\
\midrule
\multirow{2}{*}{Test}
 & DAVIS17~\cite{Ponttuset17}       & 30      & 11      & 15   & 4    & 19      \\
 & YTVOS19~\cite{Xu18}    & 507     & 246     & 109  & 152  & 261     \\
\midrule
\textbf{Total (Test)} & & \textbf{537} & \textbf{257} & \textbf{124} & \textbf{156} & \textbf{280} \\
\bottomrule
\end{tabular}}
\end{table}

\vspace{2mm}
\para{Annotation cost.}
Three annotators are employed for the entire curation process. Stage~1 (video-level filtering) takes roughly $17$ hours to check for all $5{,}001$ candidates. Stage~2 (TFA) requires approximately $170$ hours of annotation for the $1{,}467$ videos that need temporal labels. An additional $20$ hours are spent on quality verification and re-annotation, bringing the total annotation effort to approximately $207$ person-hours. All annotators are compensated above the local minimum wage.

\clearpage

\section{Ablation study}
\label{appsec:abla}
\subsection{Ablations on \method proposer}
\begin{table}[t]
\centering
\begin{minipage}[t]{0.53\textwidth}
    \vspace{0pt}
    \centering
    \small
    \setlength{\tabcolsep}{6pt}
    \captionof{table}{\textbf{Encoders in the \method proposer.}  $\pi^3$ and SAM2 are adopted as our default encoders. Throughout the proposer ablations, we report the proposer's performance directly, with the propagator skipped.}
    \label{tab:ablation_encoders}
    \vspace{2mm}
    \resizebox{\linewidth}{!}{
    \begin{tabular}{cccccc}
    \toprule
    \multirow{2}[4]{*}{\makecell{Geometric\\encoder}} & \multirow{2}[4]{*}{\makecell{Segmentation\\encoder}} & \multicolumn{3}{c}{DAVIS17-IM} \\
    \cmidrule(lr){3-5}
     & & mtIoU$\uparrow$ & $\mathcal{J}_{\text{mov}}\!\uparrow$ & \makecell{FP \\ count}$\downarrow$ \\
    \midrule
    \xmarkb & SAM2~\cite{ravi2024sam2} & $57.9$ & $74.8$ & $0.174$ \\
    VGGT~\cite{Wang25} & SAM2~\cite{ravi2024sam2} & $67.4$ & $80.4$ & $0.112$ \\
    DA3~\cite{lin2026depth} & SAM2~\cite{ravi2024sam2} & $67.7$ & $80.6$ & $0.141$ \\
    \midrule
    $\pi^3$~\cite{wang2026pi} & \xmarkb & $62.1$ & $77.1$ & \textbf{0.075} \\
    $\pi^3$~\cite{wang2026pi} & SAM3~\cite{carion2026sam} & \textbf{68.7} & \textbf{82.5} & $0.157$ \\
    \midrule
    $\pi^3$~\cite{wang2026pi} & SAM2~\cite{ravi2024sam2} & \textbf{68.7} & $82.2$ & $0.147$ \\
    \bottomrule
    \end{tabular}}
\end{minipage}
\hfill
\begin{minipage}[t]{0.44\textwidth}
    \vspace{0pt}
    \centering
    \small
    \setlength{\tabcolsep}{4pt}
    \captionof{table}{\textbf{Training dataset composition.} ``M.C.'' is short for Mannequin Challenge.}
    \label{tab:ablation_dataset}
    \vspace{1.5mm}
    \resizebox{\linewidth}{!}{
    \begin{tabular}{cccccc}
    \toprule
    \multirow{2}[2]{*}{\makecell{Syn.}} &  \multirow{2}[2]{*}{\makecell{\method\\-2K}} &  \multirow{2}[2]{*}{\makecell{M.C.}} & \multicolumn{3}{c}{DAVIS17-IM} \\
    \cmidrule(lr){4-6}
      &  &  & mtIoU$\uparrow$ & $\mathcal{J}_{\text{mov}}\!\uparrow$ & \makecell{FP count}$\downarrow$ \\
    \midrule
    \cmarkb & \xmarkb & \xmarkb & $53.0$ & $65.0$ & \textbf{0.097} \\
    \cmarkb & \cmarkb & \xmarkb & $67.4$ & $82.1$ & $0.170$ \\
    \cmarkb & \cmarkb & \cmarkb & \textbf{68.7} & \textbf{82.2} & $0.147$ \\
    \bottomrule
    \end{tabular}}

    \vspace{3.5mm}
    \setlength{\tabcolsep}{7pt}
    \captionof{table}{\textbf{Confidence-based loss.}}
    \label{tab:ablation_conf}
    \vspace{-1.5mm}
    \resizebox{\linewidth}{!}{
    \begin{tabular}{cccc}
    \toprule
    \multirow{2}[2]{*}{\makecell{Confidence\\loss}} & \multicolumn{3}{c}{DAVIS17-IM} \\
    \cmidrule(lr){2-4}
     & mtIoU$\uparrow$ & $\mathcal{J}_{\text{mov}}\!\uparrow$ & \makecell{FP count}$\downarrow$ \\
    \midrule
    \xmarkb & $65.4$ & $81.9$ & $0.192$ \\
    \cmarkb & \textbf{68.7} & \textbf{82.2} & \textbf{0.147} \\
    \bottomrule
    \end{tabular}}
\end{minipage}
\end{table}

\begin{table}[t]
\centering
\small
\begin{minipage}{0.48\linewidth}
\centering
\captionof{table}{\textbf{Input temporal window size for \method proposer.} *Our default setup takes an input temporal window of $0.5$s.}
\label{tab:ablation_window}
\vspace{1.5mm}
\setlength{\tabcolsep}{7pt}
\resizebox{\linewidth}{!}{
\begin{tabular}{cccc}
\toprule
\multirow{2}[2]{*}{\makecell{Temporal\\window (s)}} & \multicolumn{3}{c}{DAVIS17-IM} \\
\cmidrule(lr){2-4}
 & mtIoU$\uparrow$ & $\mathcal{J}_{\text{mov}}\!\uparrow$ & \makecell{FP count}$\downarrow$ \\
\midrule
$0.16$ & $67.1$ & $79.4$ & \textbf{0.130} \\
\;\,$0.50$* & \textbf{68.7} & $82.2$ & $0.147$ \\
$1.00$ & $66.7$ & \textbf{82.7} & $0.197$ \\
\bottomrule
\end{tabular}}
\end{minipage}
\hfill
\begin{minipage}{0.48\linewidth}
\centering
\caption{\textbf{Input frame gap for \method proposer.} *Our default setup takes the input frame gap of ``1\,1\,1\,1''.
}
\vspace{1.5mm}
\label{tab:ablation_frame_gap}
\setlength{\tabcolsep}{8pt}
\resizebox{\linewidth}{!}{
\begin{tabular}{cccc}
\toprule
 \multirow{2}[2]{*}{\makecell{Frame\\gap}} & \multicolumn{3}{c}{DAVIS17-IM} \\
\cmidrule(lr){2-4}
 & mtIoU$\uparrow$ & $\mathcal{J}_{\text{mov}}\!\uparrow$ & \makecell{FP count}$\downarrow$ \\
\midrule
\;\,1\,1\,1\,1* & $68.7$ & $82.2$ & \textbf{0.147}\\
2\,1\,1\,2 & $68.1$ & $82.9$ & $0.155$\\
3\,1\,1\,3 & \textbf{68.9} & \textbf{83.4} & $0.180$\\
\bottomrule
\end{tabular}}
\end{minipage}
\end{table}

\begin{table}[t]
\centering
\small
\begin{minipage}{0.48\linewidth}
\centering
\caption{\textbf{Fusing early-layer $\pi^3$ features in \method proposer.} *Our default setup does not take additional early-layer $\pi^3$ features.
}
\vspace{2mm}
\label{tab:ablation_early_feat}
\setlength{\tabcolsep}{7pt}
\resizebox{\linewidth}{!}{
\begin{tabular}{cccc}
\toprule
 \multirow{2}[2]{*}{\makecell{Early-layer \\ {\scriptsize $\pi^3$} feat.}} & \multicolumn{3}{c}{DAVIS17-IM} \\
\cmidrule(lr){2-4}
 & mtIoU$\uparrow$ & $\mathcal{J}_{\text{mov}}\!\uparrow$ & \makecell{FP count}$\downarrow$ \\
\midrule
feat. & \textbf{69.0} & \textbf{82.3} & $0.142$\\
attn. & $68.8$ & $82.1$ & \textbf{0.126}\\
\xmarkb* & $68.7$ & $82.2$ & $0.147$\\
\bottomrule
\end{tabular}}
\end{minipage}
\hfill
\begin{minipage}{0.48\linewidth}
\centering
\caption{\textbf{Mask decoder in \method proposer.} *Our default setup adopts a trainable mask decoder.
}
\vspace{2mm}
\label{tab:ablation_decoder}
\setlength{\tabcolsep}{8.5pt}
\resizebox{\linewidth}{!}{
\begin{tabular}{cccc}
\toprule
 \multirow{2}[4]{*}{\makecell{Mask\\decoder}} & \multicolumn{3}{c}{DAVIS17-IM} \\
\cmidrule(lr){2-4}
 & mtIoU$\uparrow$ & $\mathcal{J}_{\text{mov}}\!\uparrow$ & \makecell{FP \\ count}$\downarrow$ \\
\midrule
 SAM2 & $60.4$ & $78.2$ & $0.211$\\
 Trainable* & \textbf{68.7} & \textbf{82.2} & \textbf{0.147}\\
\bottomrule
\end{tabular}}
\end{minipage}
\end{table}

To probe the proposer directly, we skip the propagator and apply per-frame Hungarian matching between proposer outputs and the ground truth, reporting performance on DAVIS17-IM.

\vspace{1mm}
\para{Geometric and segmentation encoders.}
\Cref{tab:ablation_encoders} ablates the encoder choices for the \method proposer. Removing the geometric encoder destroys the model's grounding ability, producing many hallucinated false positives. Replacing $\pi^3$ with VGGT or DA3 yields a modest drop, suggesting that $\pi^3$ provides richer 4D priors that better disentangle object motion from camera motion.

Dropping the segmentation encoder, on the other hand, reduces the false-positive count but substantially degrades mask quality, causing large drops in mtIoU and $\mathcal{J}_{\text{mov}}$. Replacing SAM2 with the more recent SAM3~\cite{carion2026sam} yields comparable mask quality and only a marginal $\mathcal{J}_{\text{mov}}$ gain. We therefore retain SAM2 as the default segmentation encoder for direct comparison with prior methods that also build on it.

\vspace{1mm}
\para{Training dataset composition.}
\Cref{tab:ablation_dataset} investigates how the major subsets contribute to training. Synthetic data alone yields a strong proposer with a low FP count, but limited mtIoU and $\mathcal{J}_{\text{mov}}$ due to the synthetic-to-real domain gap. Adding the real moving-object data from \method-2K substantially improves segmentation quality, while the static-scene Mannequin Challenge clips further reduce the FP count by teaching the model to suppress false motion induced by depth parallax.

\vspace{1mm}
\para{Confidence-based loss.}
\Cref{tab:ablation_conf} shows that the confidence-based loss improves all three metrics by mitigating noisy motion annotations near static--moving transitions.

\vspace{1mm}
\para{Inference temporal window size.}
\Cref{tab:ablation_window} varies the temporal window from which five input frames are uniformly sampled at inference. Since the proposer is trained with variable window lengths, performance remains relatively stable across settings. In particular, a larger window makes moving objects easier to discover (raising $\mathcal{J}_{\text{mov}}$) but reduces temporal granularity, resulting in more false positives. A window of $\sim0.5$\,s strikes the best balance, achieving the highest mtIoU.

\vspace{1mm}
\para{Input frame gap.}
\Cref{tab:ablation_frame_gap} replaces the uniformly sampling  frame gap (\ie ``1\,1\,1\,1'') with non-uniform gaps that emphasise the centre frame. Non-uniform gaps slightly improve $\mathcal{J}_{\text{mov}}$ on moving objects but consistently raise FP count, suggesting that uniform sampling provides the most balanced motion cues for the proposer.

\vspace{1mm}
\para{Fusing early-layer $\pi^3$ features.} By default, the \method proposer takes two feature sources as input, geometric (the last layer of the $\pi^3$ encoder) and segmentation (the last layer of the SAM2 encoder). We investigate whether adding a \emph{third} source, early-layer signals from the $\pi^3$ encoder, improves the proposer. As outlined in~\Cref{tab:ablation_early_feat}, we consider two modes, both sampling early-layer $\pi^3$ signals from layers $7$, $15$, $23$, and $31$ (of $36$). In \emph{feature} mode, activations from the sampled layers are concatenated and projected. In \emph{attention} mode, cross-attention maps between the middle frame and all other frames are concatenated and projected to form the additional signal.

\Cref{tab:ablation_early_feat} shows that both variants perform on par with the default, indicating that the final-layer $\pi^3$ feature alone already suffices for motion reasoning. We therefore retain only the final-layer $\pi^3$ feature as the geometric input.

\vspace{1mm}
\para{Mask decoder.}
As detailed in~\Cref{subsec:proposer}, the \method proposer employs a trainable hypernetwork mask decoder, in which the feature map is upsampled and attended by object embeddings. 

An alternative is to discard the upsampler and predict masks directly at the bottleneck resolution ($64{\times}64$), using off-the-shelf SAM2 to recover the full-resolution mask for each object. Specifically, we randomly sample $10$ positive and $3$ negative points from the low-resolution prediction of each object as prompts to SAM2.

\Cref{tab:ablation_decoder} compares our lightweight trainable upsampling decoder against this SAM2-based decoding. The latter underperforms on all metrics, suggesting that the low-resolution prediction does not fully capture moving-object information.

\subsection{Ablations on \method propagator}
\begin{table}[t]
\centering
\small
\caption{\textbf{Propagation strategies in \method propagator.} ``Prop.~1'' and ``Prop.~2'' denote the start-frame strategy for the first and second propagation stages. \textit{forward} starts at frame~$1$, while \textit{select} starts at the anchor frame with the highest aggregate predicted IoU on confident moving proposals. ``MOS-I'' and ``MOS'' indicate which evaluation protocols each configuration supports.}
\vspace{2mm}
\label{tab:ablation_prop}
\setlength{\tabcolsep}{8pt}
\resizebox{\linewidth}{!}{
\begin{tabular}{lcccccccc}
\toprule
 \multirow{2}[2]{*}{\makecell{Exp.}} & \multirow{2}[2]{*}{\makecell{Prop. 1}} &   \multirow{2}[2]{*}{\makecell{Prop. 2}} &   \multirow{2}[2]{*}{\makecell{Online}} &   \multirow{2}[2]{*}{\makecell{MOS-I}} &   \multirow{2}[2]{*}{\makecell{MOS}} & \multicolumn{3}{c}{DAVIS17-IM} \\
\cmidrule(lr){7-9}
 & & & & & &  mtIoU$\uparrow$ & $\mathcal{J}_{\text{mov}}\!\uparrow$ & \makecell{FP count}$\downarrow$ \\
\midrule
A \textit{(online default)} & forward & - & \cmarkb & \cmarkb & \xmarkb &$72.9$ & $83.5$ & $0.075$ \\
B & forward & forward & \xmarkb & \cmarkb & \cmarkb & $72.4$ & $83.7$ & $0.080$ \\
C & select & forward & \xmarkb & \cmarkb & \cmarkb & $73.3$ & $82.9$ & $0.039$ \\
D & select & select & \xmarkb & \cmarkb & \cmarkb & $74.1$ & $84.0$ & $0.043$ \\
E \textit{(offline default)} & forward & select & \xmarkb & \cmarkb & \cmarkb & $72.9$ & $84.2$ & $0.079$\\
\bottomrule
\end{tabular}}
\vspace{-2mm}
\end{table}

\Cref{tab:ablation_prop} investigates the start-frame strategy for each propagation stage in the \method propagator. Exp.~A corresponds to our default online setup, with a single forward propagation starting at frame~$1$. Since the sequence is traversed only once in temporal order, masks cannot be recovered for every frame (\eg an object initially static but moves later). Therefore, \method (online) is restricted to the MOS-I setup. 

The four remaining variants (Exp.~B--E) add a second propagation stage, enabling both MOS-I and MOS evaluation. Among these, Exp.~C and D start the first propagation from a selected anchor frame and perform slightly better overall, but they sacrifice causality in the first propagation stage. We therefore adopt Exp.~E as our offline default, since its first stage coincides with the online procedure and its second stage further refines the result via a curated prompt set.

\begin{table}[t]
\centering
\small
\caption{\textbf{Repeated runs of \method on DAVIS17-IM.} Exp.~A--D train the proposer with four random seeds, and the deterministic propagator is then applied on top of each.}
\vspace{2mm}
\label{tab:ablation_repeat}
\setlength{\tabcolsep}{8pt}
\resizebox{0.8\linewidth}{!}{
\begin{tabular}{ccccccc}
\toprule
\multirow{2}[2]{*}{Exp.} & \multicolumn{3}{c}{DAVIS17-IM (\method proposer)} & \multicolumn{3}{c}{DAVIS17-IM (\method propagator)} \\
\cmidrule(lr){2-4}  \cmidrule(lr){5-7}
 & mtIoU$\uparrow$ & $\mathcal{J}_{\text{mov}}\!\uparrow$ & FP count$\downarrow$ & mtIoU$\uparrow$ & $\mathcal{J}_{\text{mov}}\!\uparrow$ & FP count$\downarrow$ \\
\midrule
A & $68.7$ & $82.2$ & $0.147$ & $72.9$ & $84.2$ & $0.079$ \\
B & $69.2$ & $81.9$ & $0.147$ & $72.4$ & $84.1$ & $0.077$ \\ 
C & $69.0$ & $82.6$ & $0.135$ & $73.7$ & $85.5$ & $0.045$ \\ 	
D & $68.3$ & $83.4$ & $0.133$ & $71.7$ & $85.1$ & $0.081 $\\
\midrule
Avg. & $68.8{\scriptstyle\pm0.4}$ & $82.5{\scriptstyle\pm0.7}$ & $0.141{\scriptstyle\pm0.008}$ & $72.7{\scriptstyle\pm0.8}$ & $84.7{\scriptstyle\pm0.7}$ & $0.071{\scriptstyle\pm0.017}$ \\
\bottomrule
\end{tabular}}
\vspace{2mm}
\end{table}

\subsection{Repeating experiments}
The training of \method proposer is stochastic, while the propagator step is fully deterministic. We repeat the proposer training with four different random seeds (Exp.~A--D), report the per-frame Hungarian-matched results in the left half of \Cref{tab:ablation_repeat}, and apply the deterministic propagator on top of each to obtain the sequence-level (object ID associated) results in the right half.
The small standard deviations on both halves indicate that \method is statistically stable across seeds, and the propagator brings consistent gains over the proposer alone in every run.

\section{Additional qualitative visualisations}
\label{appsec:qual}
\begin{figure*}[t]
\includegraphics[width=1.0\linewidth, trim=0 0 0 0, clip]{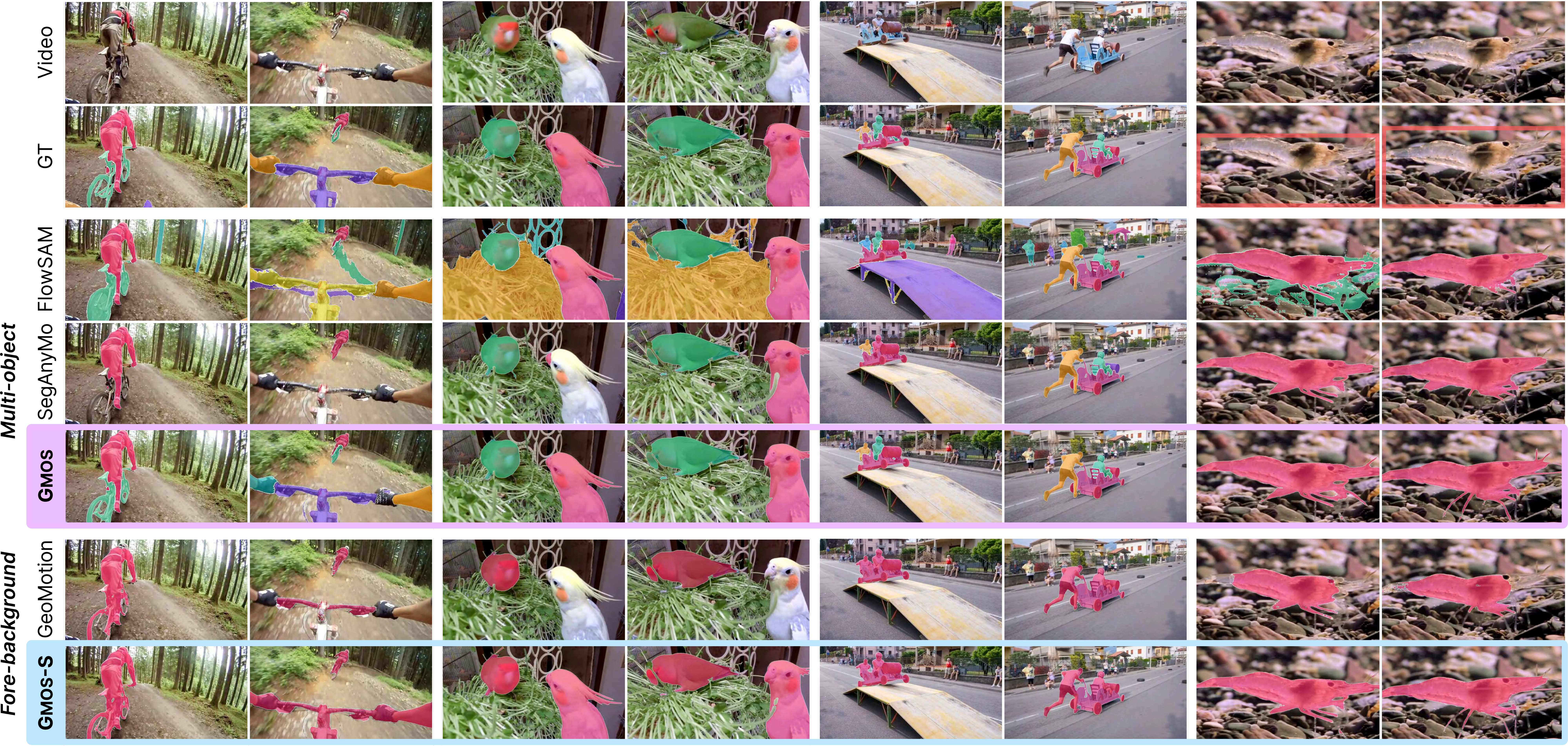}
\vspace{-4mm}
\caption{
\textbf{Additional qualitative comparison on MOS.}
Example videos are sampled from YTVOS19 (first two columns), DAVIS17 (third column), and MoCA (last column). The middle block compares the multi-object methods, while the bottom block evaluates the foreground--background models.
}
\label{fig:qual_supp}
\end{figure*}

\begin{figure*}[t]
\includegraphics[width=1.0\linewidth, trim=0 0 0 0, clip]{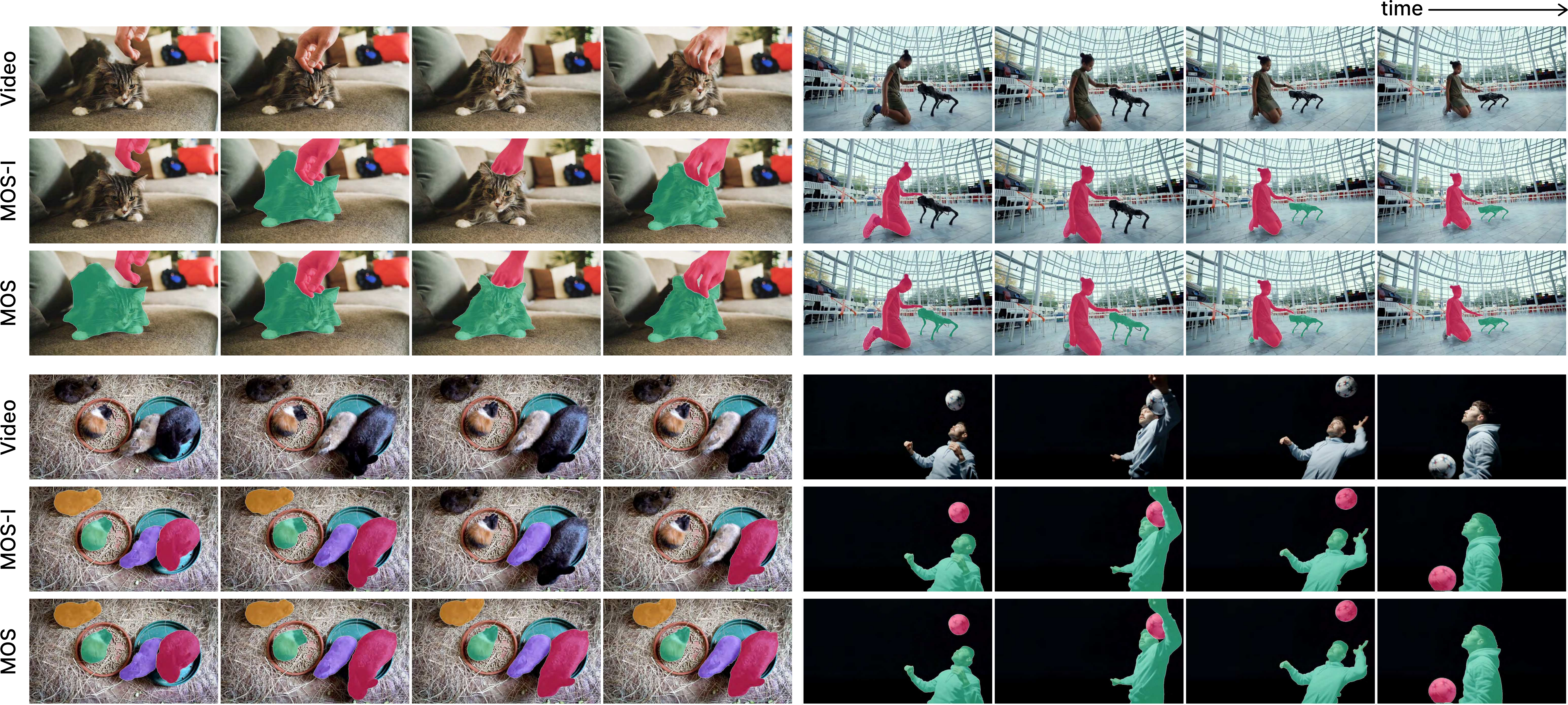}
\vspace{-4mm}
\caption{
\textbf{Additional \method results on in-the-wild videos.}
Four in-the-wild sequences (sourced outside our training or test datasets) contrast \method's MOS and MOS-I predictions. Under MOS-I, only instantaneously moving objects are segmented at each frame, with object identities consistently associated across frames. Under MOS, \method produces full segmentation masks for every object that moves at any point in the sequence.
}
\vspace{-2mm}
\label{fig:qual_supp_wild}
\end{figure*}

We provide additional qualitative results to complement \Cref{fig:qual,fig:qual_main_wild} in the main paper. \Cref{fig:qual_supp} shows further comparisons against representative baselines under the multi-object and foreground--background protocols, and \Cref{fig:qual_supp_wild} illustrates \method's behaviour on more in-the-wild sequences, distinguishing between MOS-I and MOS predictions.

\section{Discussion}
\label{appsec:disc}
\begin{figure*}[t]
\includegraphics[width=\linewidth, trim=0 0 0 0, clip]{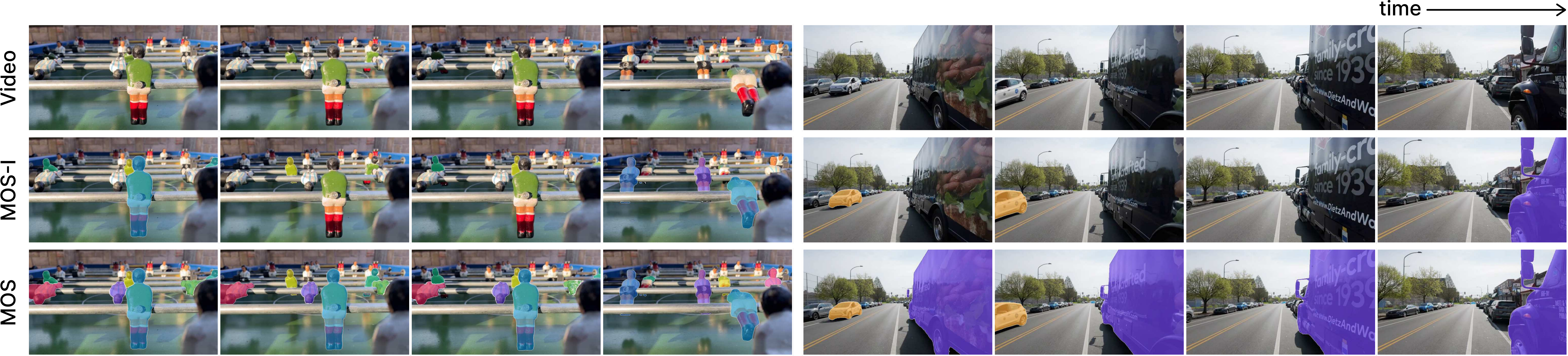}
\vspace{-4mm}
\caption{
\textbf{Failure cases of \method.} \textbf{Left:} the model struggles when many moving objects are heavily occluded. \textbf{Right:} the model occasionally produces false positives in motion-state prediction.}
\label{fig:failure_vis}
\vspace{-2mm}
\end{figure*}

\noindentpara{Failure cases.}
\Cref{fig:failure_vis} illustrates two representative failure modes, both stemming from the limited spatial granularity of the $\pi^3$ geometric backbone. The left example (a table football game) shows a scene with many fast-moving objects under heavy occlusion. \method misses several small moving instances in the background, suggesting that the backbone struggles with fine-grained moving-object discovery in cluttered scenes. The right example shows a driving scene in which a large reflective truck passes the camera. Although the truck is static, \method incorrectly predicts it as moving at a particular frame, exposing limitations of the backbone in per-object motion-state prediction under reflections and extreme depth parallax.

\vspace{2mm}
\para{Possible extensions.}
We outline two natural directions for future work. First, the geometric backbone could be fine-tuned on larger and more diverse motion-annotated datasets to strengthen both fine-grained moving-object discovery and motion-state prediction. Second, the proposer and propagator could be merged into a single end-to-end trainable framework, allowing the temporal consistency signal to improve mask quality and yield more accurate motion-state predictions over long videos.

\vspace{2mm}
\para{Broader impact.}
\method enables real-time identification of moving objects directly from RGB video, supporting safety-critical applications such as autonomous driving, ecological and biomedical monitoring, public-safety surveillance, and dynamic 3D/4D scene reconstruction for robotics and AR/VR. Operating without pre-computed flow or trajectories also lowers the compute footprint of motion-aware perception. We release \method-2K and the MOS-I protocol for reproducible research.
As with any video-segmentation system, \method could be misused in surveillance contexts that raise privacy concerns, particularly when combined with identity-recognition tools. We encourage downstream users to comply with data-protection regulations and obtain appropriate consent.

\end{document}